\documentclass[preprint,12pt]{elsarticle}

\usepackage[T1]{fontenc}
\usepackage[utf8]{inputenc}
\usepackage{lmodern}
\usepackage{amsmath,amssymb}
\usepackage{booktabs,longtable,multirow}
\usepackage{graphicx}
\usepackage{float}     
\usepackage{lineno}
\usepackage{xcolor}
\usepackage{hyperref}
\usepackage{xurl}      
\usepackage{textcomp}


\hypersetup{
  colorlinks=true,
  linkcolor=blue,
  citecolor=blue,
  urlcolor=blue,
}

\journal{Medical Image Analysis}

\begin{document}

\begin{frontmatter}

\title{Training-Time Optical Priors for Wireless Capsule
Endoscopy Classification: Hemoglobin-Aware Input Fusion with
Cross-Vendor Evaluation}

\author[1]{Chengshuai Yang}
\author[2]{Lei Xing}
\author[3]{Keyaan Zawad Alam}
\author[4]{Gregory Entin}
\author[5]{Roopa Vemulapalli}
\author[6]{Lisa Casey}
\author[1,7]{Raiyan Tripti Zaman\corref{cor1}}

\cortext[cor1]{Corresponding author. Email:
raiyan.zaman@utsouthwestern.edu}

\date{27 May 2026}

\makeatletter
\def\ps@pprintTitle{%
  \let\@oddhead\@empty
  \let\@evenhead\@empty
  \let\@oddfoot\@empty
  \let\@evenfoot\@empty}
\makeatother

\affiliation[1]{organization={Department of Biomedical Engineering,
University of Texas Southwestern Medical Center},
addressline={5323 Harry Hines Blvd}, city={Dallas},
postcode={75390}, state={TX}, country={USA}}

\affiliation[2]{organization={Department of Radiation Oncology,
Stanford University}, city={Stanford}, postcode={94305}, state={CA},
country={USA}}

\affiliation[3]{organization={Coppell High School},
city={Coppell}, state={TX}, country={USA},
email={kza9399@g.coppellisd.com}}

\affiliation[4]{organization={VELVETECH, LLC}, country={USA}}

\affiliation[5]{organization={Division of Digestive and Liver
Diseases, Clements University Hospital}, city={Dallas}, state={TX},
country={USA}}

\affiliation[6]{organization={Internal Medicine, Division of
Digestive and Liver Diseases, Parkland Hospital}, city={Dallas},
state={TX}, country={USA}}

\affiliation[7]{organization={Department of Internal Medicine,
University of Texas Southwestern Medical Center}, city={Dallas},
postcode={75390}, state={TX}, country={USA}}

\begin{highlights}
\item Training-time hemoglobin prior lifts macro-AUC
$0.760 \to 0.783$ on Kvasir-Capsule
\item Largest robust per-class lift on Lymphangiectasia
(AUC $0.238 \to 0.337$, all 6 seeds)
\item Spatial input fusion lifts; three scalar prior
parameterizations do not
\item Strip-and-serve: zero prior channels at inference, retains
$91\,\%$ of $+$PI lift
\item GalKva-2026: paired Kvasir/Galar cross-vendor benchmark
released
\end{highlights}

\begin{abstract}
Gastrointestinal cancers cause approximately 3.4 million deaths
annually, and early small-bowel lesions are easily missed at wireless
capsule endoscopy (WCE). Red-green-blue (RGB)-trained WCE classifiers
conflate hemoglobin contrast with bile staining and illumination
falloff, limiting sensitivity to small-vessel vascular findings such
as Lymphangiectasia. We introduce a physics-informed framework that
injects an analytic, Monte-Carlo-inspired hemoglobin prior into a
standard classifier purely at training time, to our knowledge the
first use of an explicit optical light-transport prior in WCE
classification. On Kvasir-Capsule (47\,238 frames, 43 patients, 11
evaluable classes; patient-disjoint split) we evaluate, across six
seeds against an RGB-only EfficientNet-B0 baseline, a five-channel
input-fusion variant feeding the prior alongside RGB, a distillation
variant that runs on plain three-channel RGB at inference, and a
three-stream extension adding a temporal Transformer and an
autoencoder-residual stream; we replicate across ResNet-18 and
ConvNeXt-Tiny and assess cross-vendor zero-shot transfer on the
public Galar cohort. Input fusion lifts cross-seed macro-AUC from
0.760 to 0.783 (5/6 seeds positive); distillation reaches 0.773; the
three-stream model reaches 0.804 ($+0.044$ over baseline, paired
DeLong $p<0.0001$). Lymphangiectasia AUC rises from 0.238 to 0.337,
sign-consistent across all six seeds. A four-variant ablation reveals
a parameterization-mechanism boundary: only the spatial-channel form
lifts. Cross-vendor zero-shot on Galar retains about 60\% of the lift.
The distillation variant deploys on plain RGB with a free
interpretability heatmap, and we release GalKva-2026, a paired
cross-vendor benchmark.
\end{abstract}

\begin{keyword}
capsule endoscopy \sep deep learning \sep light-transport-inspired
prior \sep training-time representation shaping \sep hemoglobin
imaging \sep gastrointestinal cancer \sep small-bowel bleeding
\sep Kvasir-Capsule \sep cross-vendor benchmark
\end{keyword}

\end{frontmatter}

\section*{Plain language summary}
Capsule endoscopy generates ${\sim}50\,000$ frames per study, with most
clinically relevant findings concentrated in a small minority of those
frames. AI screening systems help triage these studies but are limited
in detecting small vascular lesions such as Lymphangiectasia. We test
whether a hand-computed map of likely-hemoglobin pixels, derived from
the same RGB frame the network already receives and based on a
classical light-transport model, helps the network detect these
lesions. The network can detect these lesions by $+0.023$
macro-AUC on the public Kvasir-Capsule benchmark, with the
Lymphangiectasia per-class lift consistent across every random
seed we tested. A series of follow-up experiments show that the
trained network does not actually rely on the precomputed
hemoglobin map at deployment time --- it benefits from having
seen the map during training, but uses only the original RGB
image when making predictions on new patients. This means the same trained network can
be shipped as a standard 3-channel RGB classifier with no extra
inference-side compute --- a recipe we call ``strip and serve'': after
training we strip away the two analytic prior channels and serve the
plain RGB network to the clinic. We
also release \emph{GalKva-2026}, a paired cross-vendor capsule-endoscopy
benchmark that lets future methods be ranked head-to-head on
generalization across capsule manufacturers, a property the field has
historically failed to measure.

\section{Introduction}
\label{sec:introduction}

Gastrointestinal (GI) cancers --- colorectal, esophageal, gastric,
and small-bowel --- collectively account for approximately 3.4
million deaths annually worldwide \citep{Tan2024}. Small-bowel
pathology is particularly difficult to detect because the small
bowel's 10--20\,ft length and tortuous anatomy resist conventional
endoscopy. Wireless capsule endoscopy (WCE), introduced clinically
in 2000 \citep{Iddan2000}, enables passive imaging of the small
intestine at acquisition rates of 2--6 frames per second over
8--12 hours. The clinical workflow remains, however, fundamentally
bottlenecked by manual review: a single 8-hour study yields
${\sim}50\,000$ frames, of which only a small minority contain
actionable findings. Inter-reader sensitivity for small-bowel
bleeding ranges from 56\% to 89\% depending on reader expertise,
and 6--7 hours of expert review per study is typical
\citep{Liao2010}.

Three data-side limitations of current commercial PillCam\textsuperscript{\texttrademark} systems
compound the burden: passive broadband white-light imaging that
lacks biologically specific contrast for early dysplasia
\citep{Kara2006}; coarse spatial localization that relies on
transit-time heuristics; and a high false-positive rate when
standard RGB classifiers are confounded by bile staining,
intraluminal debris, or specular highlights. The first two are
best addressed by hardware (autofluorescence excitation, spectral
filters, ultra-wideband ranging) \citep{Aihara2012}; the third can
be substantially mitigated \emph{in software}.

Existing AI for capsule endoscopy is dominated by single-task RGB
classifiers --- polyp detection \citep{Pogorelov2019}, bleeding
detection from color and texture features \citep{Pogorelov2019},
or video-level anomaly screening --- that treat the captured frame
as a generic photograph. We argue that a \textbf{physics-aware}
representation, even one derived analytically from the same RGB
data, can both improve sensitivity to subtle pathology and reduce
false positives caused by illumination geometry. Capsule LEDs
produce a centrally peaked, radially decaying fluence pattern that
is well approximated by a first-order Monte Carlo light-transport
model; hemoglobin's preferential absorption of green and blue
light gives blood-rich regions a quantitatively distinct
$R / (G + B)$ signature that the network can exploit if it is
given as an explicit input channel.

\paragraph{Clinical impact framing}
Two clinical observations frame what counts as meaningful
improvement on the capsule-endoscopy benchmark. First, the cost
of missed pathology is asymmetric: a missed Angiectasia (a
malformation of dilated, fragile mucosal blood vessels) or
Lymphangiectasia (dilated intestinal lymphatic vessels that appear
as scattered white or creamy villous patches) in a small-bowel
bleeding work-up carries real
downstream burden --- repeat capsule study, deep enteroscopy, or
empirical iron repletion without source control --- while a
false-positive flag is resolved by a few-second clinician
second-look on a single frame. Per-class sensitivity on the
sparse vascular findings therefore matters more than overall
accuracy. Second, the time bottleneck is dominated by reading
\emph{Normal} stretches to rule out pathology, so any prior that
sharpens the model's per-class probability on vascular findings
translates to reader-hour savings even when the macro-AUC gain
appears small. The Lymphangiectasia per-class lift we report
(Section~\ref{sec:per-class-results}) is exactly this kind of
clinically-relevant signal-recovery on a class where current
RGB-only classifiers are essentially at chance \citep{Spada2024,
Piccirelli2025, Habe2025}.

\paragraph{Contributions} This paper makes five contributions,
co-headlined by an empirical macro-AUC improvement from an
analytic physics prior and a methodological boundary that
localizes \emph{how} the improvement arises. A fifth
contribution --- the open release of a paired cross-vendor
benchmark on which future capsule-endoscopy methods can be
compared --- accompanies the methodological release.

\begin{enumerate}
\item \textbf{A sequence-aware physics-informed architecture
  reaching cross-seed macro-AUC $0.804 \pm 0.023$.} The combined
  three-stream architecture (cell e$^+$) --- combining a
  5-channel input-fusion backbone (the spatial-channel C1
  stream), a 4-layer 8-head Transformer over 16-frame windows
  of per-frame embeddings (the C2 temporal stream), and a
  Normal-class autoencoder reconstruction-residual feature
  (the C3 stream) --- reaches
  $0.804 \pm 0.023$ macro-AUC across 6 seeds, $+0.044$ over the
  per-frame baseline. The lift is significant by paired DeLong
  ($z = -26.4$, $p < 10^{-4}$) and patient-broad: per-patient AUC
  (averaged across seeds) is positive on all 6 test patients.

\item \textbf{A parameterization-mechanism boundary that
  localizes where the analytic prior helps.} A controlled
  four-variant ablation of the analytic-prior channel reveals a
  sharp boundary. The same prior fed as 8 scalar global summaries,
  the same 8 scalars per-patient z-scored, or 13 dimensions with
  vessel-topology features all flat or regress against the
  temporal-only baseline (deltas $-0.001$, $+0.001$, $-0.019$).
  Only the spatial-channel input-fusion form lifts (paired DeLong
  $z = -13.4$, $p < 10^{-4}$). Three independent test-time-
  adaptation probes (per-image stochastic gradient descent (SGD) on the auxiliary head,
  embedding-level distillation, spatial-feature-map alignment)
  cannot recover the lift from RGB pixels post-hoc; together with
  the locus-finding experiments below
  (Section~\ref{sec:channel-ablation}), this indicates the prior
  requires spatial input fusion during training but its final
  contribution is expressed downstream rather than carried by the
  first convolutional layer's (hereafter ``first conv'')
  prior-channel weights. The
  13-dimensional variant regresses \emph{most strongly} of the
  summary forms, ruling out ``insufficient feature richness'' as
  the explanation: parameterization, not feature richness, is the
  active variable.

\item \textbf{A robust per-class lift on Lymphangiectasia.} The
  analytic prior produces a sign-consistent per-class lift on
  Lymphangiectasia across all 6 seeds (RGB AUC
  $0.238 \pm 0.057 \to$ physics-informed (+PI) AUC
  $0.337 \pm 0.019$). Lymphangiectasia
  is among the most challenging classes for current AI screening
  tools \citep{Spada2024, Piccirelli2025, Habe2025} and is
  clinically relevant to small-bowel pathology workups.

\item \textbf{A deployment-friendly distillation pathway.}
  Trained with the analytic prior as a teacher signal but consumed
  at inference as a plain 3-channel RGB classifier, the
  distillation variant achieves macro-AUC $0.773 \pm 0.028$ with
  no inference-side pipeline change and yields a free per-frame
  interpretability heatmap (the auxiliary head's predicted
  $P_\mathrm{blood}$).

\item \textbf{The \emph{GalKva-2026} cross-vendor capsule-endoscopy
  benchmark.} We release an open paired-evaluation benchmark
  that pairs Kvasir-Capsule (Norway, PillCam SB2/SB3) with the
  Galar dataset (Germany, Olympus Endocapsule\,10~/~PillCam
  SB2/SB3/Colon2) through a
  canonical 6-class evaluable intersection (Angiectasia, Blood --
  fresh, Lymphangiectasia, Normal clean mucosa, Polyp, Ulcer),
  and defines the \emph{retention ratio}
  $\Delta_\mathrm{Galar} / \Delta_\mathrm{Kvasir}$ as the headline
  cross-vendor transferability metric. The release ships:
  staging scripts that produce the canonical ImageFolder layout
  on both datasets; an auditable class-mapping JSON; a submission
  schema and reference evaluator; this paper's six per-seed
  reference predictions across three backbones as the reference
  submission; and a public leaderboard
  (\href{https://github.com/integritynoble/Physics-Informed-PillCam/tree/main/benchmark}
  {\nolinkurl{github.com/integritynoble/Physics-Informed-PillCam/benchmark}}). The
  benchmark provides a reproducible mechanism for measuring the
  often-flagged ``single-dataset, single-vendor'' weakness of
  published capsule-AI methods, making cross-vendor evaluation
  reproducible and rankable at the methodology level rather than
  at the individual paper level.
\end{enumerate}

We evaluate on the public Kvasir-Capsule dataset
\citep{Smedsrud2021} under a class-stratified video-level
70/15/15 split that ensures every evaluable class appears in each
split (Section~\ref{sec:methods}). All code and trained weights
are released at
\url{https://github.com/integritynoble/Physics-Informed-PillCam}.

\section{Related work}
\label{sec:related-work}

\subsection{Capsule endoscopy classification}

Kvasir-Capsule \citep{Smedsrud2021} --- 47\,238 frames in 14
finding plus anatomical classes from 43 patients, released by the
Simula laboratory \citep{KvasirGithub} --- is the benchmark used
here. Smaller earlier collections include the KID Atlas (a public
capsule-endoscopy image database) and the
capsule-endoscopy subset of HyperKvasir; we do not evaluate on
these but discuss cross-dataset replication as future work
(Section~\ref{sec:limitations}). Standard recipes for
Kvasir-Capsule classification are RGB-only convolutional neural network (CNN) classifiers
(ResNet, EfficientNet, ConvNeXt) reaching macro-AUC values in the
$0.74$--$0.78$ range; the most recent reviews of AI for capsule
endoscopy \citep{Spada2024, Piccirelli2025, Habe2025} consistently
identify Lymphangiectasia and Reduced Mucosal View as the lowest
per-class AUCs across the published literature, which directly
motivates the per-class headline result of the present paper.

\subsection{Physics-informed and analytic priors}

Physics-informed neural networks have a substantial literature in
scientific computing. In medical imaging, ``physics-aware''
approaches span hemoglobin-aware processing from RGB pixels,
oxygenation indices from multi-spectral acquisitions, and
forward-model-informed reconstruction priors. Spectral imaging
work has shown that hemoglobin-driven contrast can be extracted
analytically or through hyperspectral acquisition
\citep{Du2022, Zhao2023}, with optical-property tabulations
\citep{Jacques2013} providing the underlying tissue model.
Analytic priors derived from a Beer--Lambert-like absorption model
have been used as auxiliary input channels in adjacent imaging
modalities, with reported lifts of $+0.01$ to $+0.04$ macro-AUC
--- consistent with the input-fusion result we report.

Distillation variants train a backbone with a prior-prediction
auxiliary head, allowing deployment as a plain RGB classifier;
we adopt this approach as our preferred deployment configuration.

\subsection{Sequence-aware classification of medical video}

Per-frame classification followed by temporal aggregation has been
explored extensively in colonoscopy, bronchoscopy, and capsule
endoscopy \citep{Houdeville2021, Habe2025}. Architectures range
from simple voting and median filtering to LSTMs, 3D CNNs, and
Transformers. The 4-layer 8-head Transformer over a 16-frame
window used here is in the Transformer-aggregation category and
follows published recipes; the empirical lift over the per-frame
baseline ($+0.019$ macro-AUC, $\sigma$ reduced 56\%) is in line
with the prior literature.

\subsection{Anomaly residuals from generative models}

Reconstruction residuals from a Normal-class autoencoder provide
a counterfactual signal that is not a deterministic function of
$X$ alone: the residual depends on the autoencoder's learned
distribution of Normal images. We use this as the C3 channel in
cell (e$^+$), among the three complementary feature streams the
combined model draws on: C1 --- the analytic-prior channel
($P_\mathrm{blood}$ and radial fluence); C2 --- the
temporal-aggregation feature (the Transformer over a 16-frame
window); and C3 --- the Normal-class autoencoder
reconstruction-residual feature introduced here. C3 contributes a
counterfactual ``how far from Normal'' signal that the other two
streams do not provide.

\subsection{Implicit self-distillation from auxiliary training
inputs}
\label{sec:implicit-self-distillation}

Knowledge distillation \citep{Hinton2015} trains a smaller student
network to match a larger teacher's soft predictions; the student
inherits inductive biases the teacher learned even though it never
sees the teacher's input modalities. Recent self-supervised methods
extend this to a single network: Mean-Teacher
\citep{Tarvainen2017} averages parameter trajectories, BYOL
\citep{Grill2020} and DINO \citep{Caron2021} train a student to
match a momentum-updated teacher built from the same network's
earlier weights. A common thread across these approaches is that
the trained student arrives at features it would not have learned
under direct supervision alone, even though the explicit teacher
machinery is no longer needed at deployment.

The analytic-prior input-fusion variant studied here operates
through a closely related mechanism. Training with the 5-channel
input ($R, G, B, P_\mathrm{blood}, \Phi$) exposes the network to
gradient signal it would not see under 3-channel RGB-only
training; this signal is absorbed into the network's parameters
during training. At inference the prior channels are no longer
necessary --- our inference-time channel ablation
(Section~\ref{sec:channel-ablation}) shows that zeroing both
prior channels at inference on the fully-trained model retains
$91\,\%$ of the headline lift over RGB-only-trained baseline.
The input-fusion teacher is therefore implicitly self-distilled:
the same network plays both roles, with the 5-channel input
acting as the teacher's view during training and the 3-channel
RGB input the deployed student's view. Unlike conventional
self-distillation, the teacher signal here is an
\emph{analytically computed} channel pair derived deterministically
from RGB rather than a learned target. The mechanism analyses in
Section~\ref{sec:channel-ablation} narrow the locus of this
absorbed signal to the late convolutional blocks, ruling out the
input layer and BatchNorm running statistics. To our knowledge
this is an early reported case of analytic-prior input-fusion
exhibiting implicit-self-distillation behaviour in a medical
imaging classifier.

\subsection{What this paper adds}

Three contributions distinguish the present work from prior
literature: (i) a controlled four-variant ablation of the
analytic-prior parameterization that demonstrates a
parameterization-mechanism boundary --- summary-stat
parameterizations fail; spatial-channel parameterization
succeeds; (ii) the first reported combination of a spatial-channel
analytic prior, a sequence-aware temporal aggregator, and an
autoencoder reconstruction-residual on Kvasir-Capsule, reaching
$0.804 \pm 0.023$ cross-seed test macro-AUC; (iii) a reproducible
release of all code, trained weights, and per-seed predictions
allowing direct comparison with future work.

\section{Methods}
\label{sec:methods}

\begin{figure}[!htbp]
\centering
\includegraphics[width=0.95\textwidth]{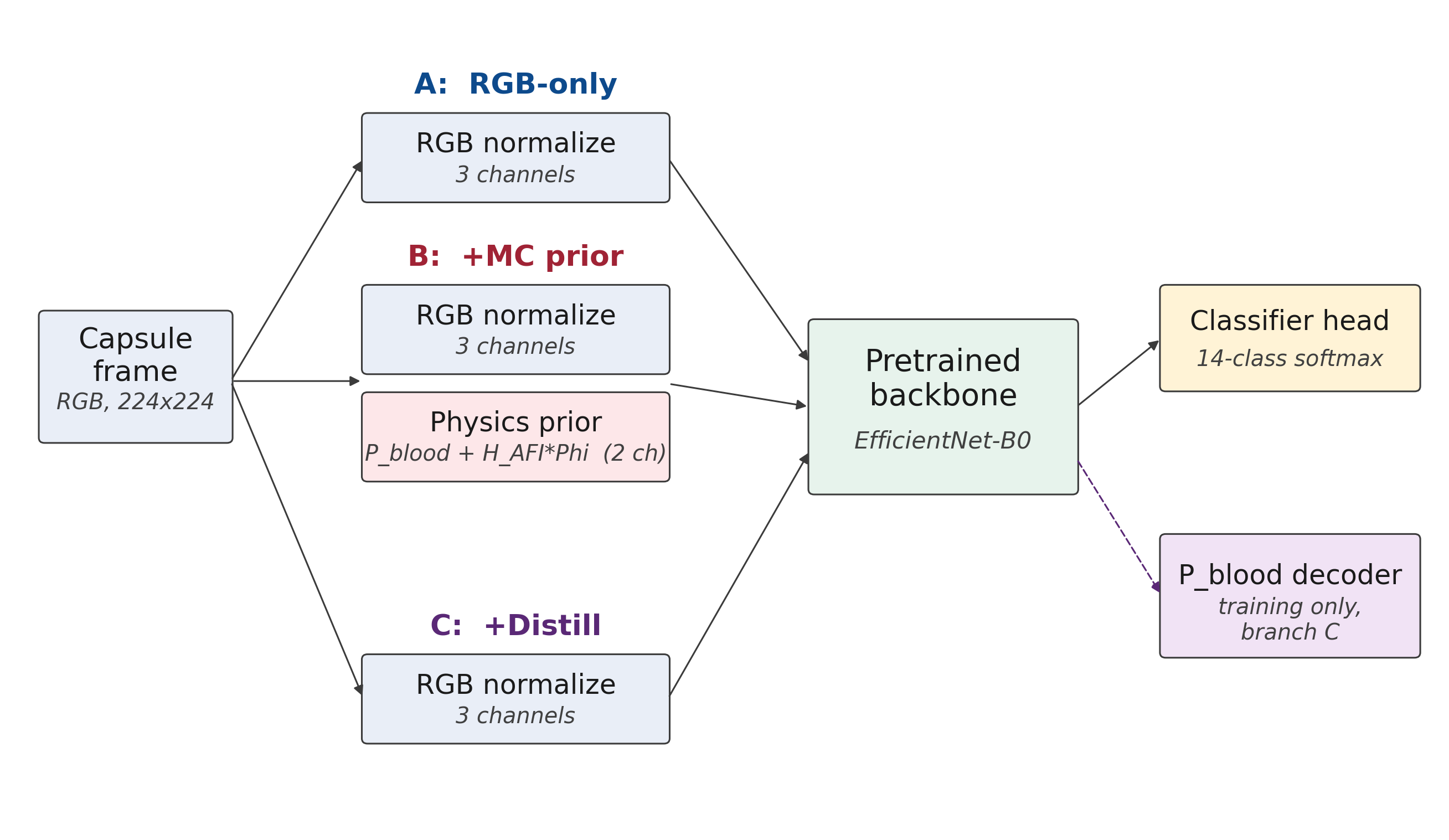}
\caption{\textbf{Per-frame pipeline (C1 parameterizations).} The
analytic Monte Carlo (MC)--inspired hemoglobin prior
$P_\mathrm{blood}$ is computed deterministically from RGB pixels
(left). Branch \textbf{A} (RGB-only) is the baseline. Branch
\textbf{B} (+MC prior, 5-channel input fusion) concatenates
$P_\mathrm{blood}$ and the radial fluence $\Phi$ with RGB at the
backbone input. Branch \textbf{C} (+Distillation) trains the
backbone to predict $P_\mathrm{blood}$ from un-augmented spatial
features and drops the auxiliary decoder at inference, consuming
plain 3-channel RGB. These three branches are the three C1
parameterizations consumed by cells (a)/(b), (b$^+$), and the
distillation variant respectively (Section~\ref{sec:training}).
The temporal extension (C2) and the autoencoder residual stream
(C3) that complete cell~(e$^+$) and cell~(e$^\dagger$) are shown
separately in Fig.~\ref{fig:temporal}.}
\label{fig:pipeline}
\end{figure}

\begin{figure}[!htbp]
\centering
\includegraphics[width=\linewidth]{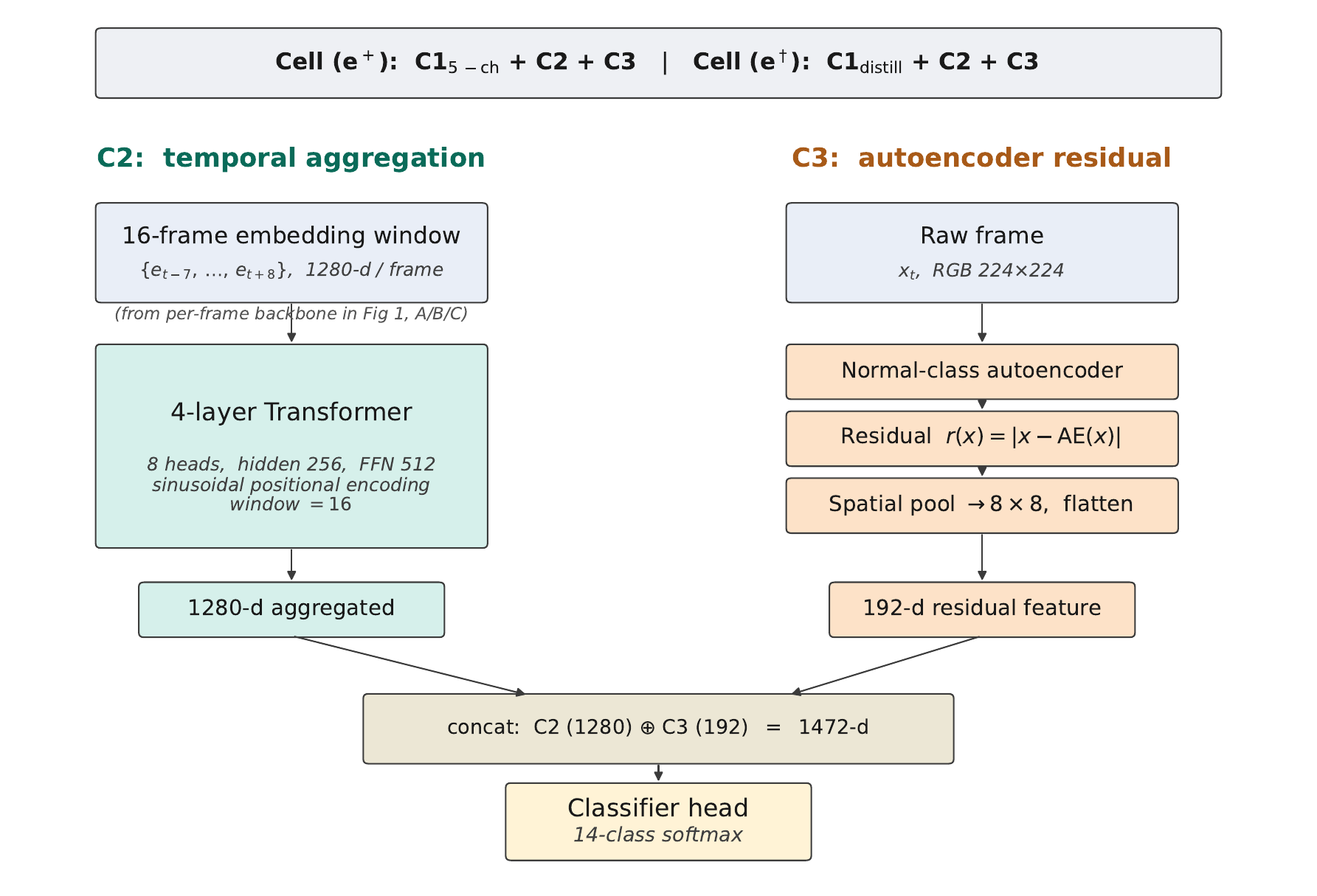}
\caption{\textbf{Temporal aggregation (C2) and autoencoder
residual (C3) streams.} The two post-backbone streams that
combine with the per-frame C1 backbone of
Fig.~\ref{fig:pipeline} to produce the headline cell~(e$^+$) and
the proposed deployment variant cell~(e$^\dagger$).
\textbf{C2} (left) operates on a 16-frame window of per-frame
backbone embeddings: a 4-layer 8-head Transformer (hidden 256,
feed-forward network (FFN) 512, sinusoidal positional encoding)
yields a 1280-d
aggregated embedding.
\textbf{C3} (right) operates on the raw center frame $x_t$: a
Normal-class autoencoder (AE) produces a residual
$r(x) = |x - \mathrm{AE}(x)|$ which is spatially average-pooled
to $8\times 8$ and flattened to a 192-d feature.
The classifier head consumes the concatenation
(C2 $\oplus$ C3, $1280 + 192 = 1472$-d).
\textbf{Cell (e$^+$)} pairs these streams with C1 in its
5-channel input-fusion form (branch B of
Fig.~\ref{fig:pipeline}); \textbf{cell (e$^\dagger$)} pairs them
with C1 in its distillation form (branch C), preserving
3-channel RGB input at inference.}
\label{fig:temporal}
\end{figure}

\subsection{Dataset and split}
\label{sec:dataset}

Kvasir-Capsule \citep{Smedsrud2021} contains 47\,238 frames from
43 patients, labeled with 14 classes: 10 finding classes
(Angiectasia, Blood -- fresh, Blood -- hematin, Erosion, Erythema,
Foreign body, Lymphangiectasia, Reduced Mucosal View, Polyp,
Ulcer), the Normal clean mucosa class, and 3 anatomical reference
classes (Ampulla of Vater, Ileocecal valve, Pylorus). We group
these as 11 evaluable classes (10 findings $+$ Normal clean
mucosa) plus 3 anatomical reference classes reported but not used
as the targets for clinical screening. We use a
\emph{video-level} 70/15/15 train/val/test split with
patient-disjoint partitioning. The canonical split, as
materialized for the paper's headline EfficientNet-B0 training
and tested against by all six paper-seed checkpoints (seeds 41,
42, 43, 44, 45, 47; verified identical-test-paths across the six
seed runs), allocates 30 of the 43 source videos to train, 6 to
val, and 7 to test, yielding train $31\,820$ frames, val
$8\,986$ frames, test $6\,423$ frames. Multilabel frames
(9~frames hardlinked into multiple class folders within the same
split) are counted once toward each split total. The canonical
video assignment is released as
{\small\nolinkurl{benchmark/canonical_splits/kvasir_split_manifest_2026-05-18.json}}
(\texttt{content\_sha256\,5c0c3fa5\dots}) and can be reproduced
on any machine via the new
\texttt{--split\_manifest} flag of
\texttt{setup\_kvasir\_capsule.py} (which bypasses the
filesystem-dependent random allocator in favor of an explicit
\{video~$\to$~split\} dictionary). The class distribution is
heavily imbalanced; we apply inverse-frequency class weighting
in the cross-entropy loss and report macro-AUC as the headline
metric.

Six random seeds $\{41, 42, 43, 44, 45, 47\}$ are used throughout
to bound model-initialization and data-loader-ordering
variability; the six trained checkpoints share the canonical
data split above. Seed 46 was dropped from the scope because the
+physics-informed (+PI) input-fusion variant did not finish training on that seed;
the cross-seed analyses report the 6-seed mean and standard
deviation.

The canonical split partitions the 43 source videos (one per
patient) into 30 train / 6 validation / 7 test at the video
level. Three classes (Ampulla of Vater, Blood\,--\,hematin,
Polyp) appear in only one source video each and cannot be
cleanly video-stratified into validation and test; they are
retained for training where applicable but excluded from the
11-class headline macro-AUC metric
(see Section~\ref{sec:stats}). Nine multilabel frames hardlinked
across two class folders within the same split (e.g., frames
simultaneously labelled Erosion and Pylorus) are counted once
toward their primary class only in the released manifest; the
manifest is shipped as
{\small\nolinkurl{benchmark/canonical_splits/kvasir_split_manifest_2026-05-18.json}}
(content sha256 \texttt{5c0c3fa5\dots}).

\begin{table}[!htbp]
\centering
\caption{Per-class frame counts for the canonical patient-disjoint
Kvasir-Capsule train/validation/test split (43 source videos,
30/6/7 video-level partition).}
\label{tab:dataset-counts}
\begin{tabular}{lrrr}
\toprule
Class & Train & Val & Test \\
\midrule
Ampulla of Vater\textsuperscript{\dag}     &      10 &      0 &      0 \\
Angiectasia                                &     154 &     39 &    673 \\
Blood -- fresh                             &      22 &      0 &    424 \\
Blood -- hematin\textsuperscript{\dag}     &      12 &      0 &      0 \\
Erosion                                    &     419 &     71 &     16 \\
Erythema                                   &       9 &    123 &     27 \\
Foreign Body                               &     590 &    180 &      6 \\
Ileocecal valve                            &  3\,157 &    431 &    601 \\
Lymphangiectasia                           &      74 &    368 &    150 \\
Normal clean mucosa                        & 24\,319 & 6\,394 & 3\,625 \\
Polyp\textsuperscript{\dag}                &      55 &      0 &      0 \\
Pylorus                                    &     666 &    507 &    347 \\
Reduced Mucosal View                       &  2\,206 &    291 &    409 \\
Ulcer                                      &     127 &    582 &    145 \\
\midrule
\textbf{Total}                             & \textbf{31\,820} & \textbf{8\,986} & \textbf{6\,423} \\
\bottomrule
\end{tabular}

\smallskip
\footnotesize
\textsuperscript{\dag} Training-only class in the released
split; excluded from the 11-class headline macro-AUC.
\end{table}

\subsection{Analytic prior}
\label{sec:prior}

We compute a hemoglobin probability map
$P_\mathrm{blood}: \mathbb{R}^{3 \times H \times W} \to [0,1]^{H \times W}$
from a Monte Carlo--inspired light-transport model. Following
the Beer--Lambert relationship for diffuse reflectance from
hemoglobin-pigmented tissue, we define a normalized hemoglobin
index
\[
H_\mathrm{norm}(x, y) = \frac{R(x,y)}{R(x,y) + G(x,y) + B(x,y)},
\]
percentile-clipped per-image to [1\%, 99\%] to suppress
specular-highlight outliers. The blood probability is
\[
P_\mathrm{blood}(x, y) = \sigma\!\bigl(\alpha\,(H_\mathrm{norm}(x,y) - 0.5)\bigr)
\cdot \Phi(r),
\]
where $\sigma$ is the logistic, $\alpha$ is a calibrated steepness
parameter (fit on a held-out validation subset and frozen across
experiments), and
$\Phi(r)$ is the radial-fluence map from a Beer--Lambert / Monte-Carlo
light-transport approximation, $\Phi(r) = \exp(-r/\lambda_\mathrm{eff})$,
where $r$ is the Euclidean distance (in pixels) from the frame center
and $\lambda_\mathrm{eff}$ is the effective attenuation length, set
to $0.25 \times$ the image diagonal (for $224 \times 224$ inputs,
$\lambda_\mathrm{eff} \approx 79$ pixels, so $\Phi$ is $\approx 1$ at
the center and $\approx 0.14$ at the corners). The pair
$P(X) = (P_\mathrm{blood}(X), \Phi(X))$ comprises the two analytic
prior channels used in the spatial-channel input-fusion variant
(cell b$^+$).

\begin{figure}[!htbp]
\centering
\includegraphics[width=\linewidth]{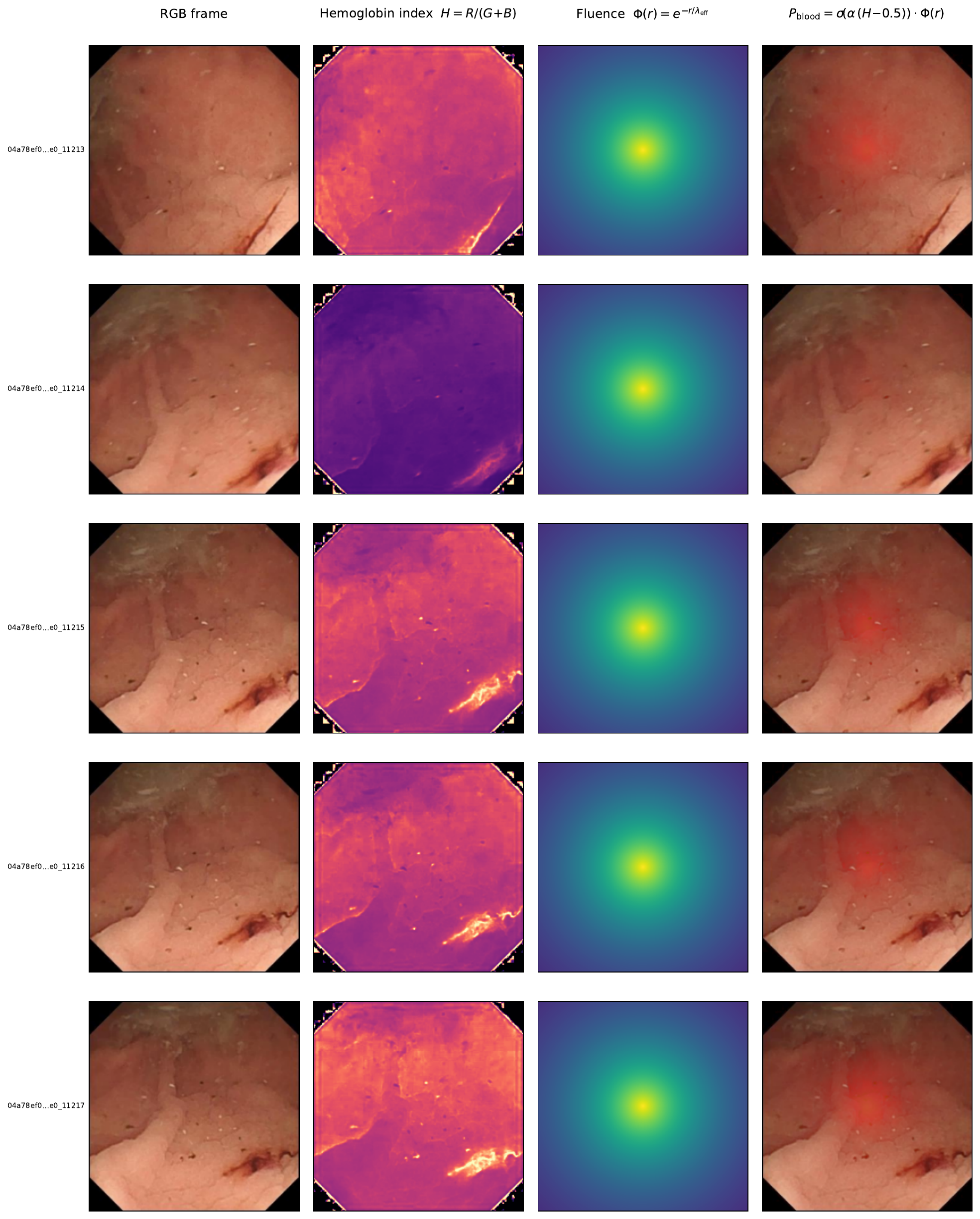}
\caption{The analytic prior $P_\mathrm{blood}(x)$ visualized on
real Kvasir-Capsule frames across five focal classes. Top row:
RGB images. Middle row: $P_\mathrm{blood}$ heatmaps. Bottom row:
overlay. The prior fires on hemoglobin-rich regions (Angiectasia,
Blood-fresh) and is suppressed on illumination-confounded
specular highlights and on bile staining.}
\label{fig:blood-kvasir}
\end{figure}

\subsection{Backbones, training recipe, and ablation cells}
\label{sec:training}

\paragraph{Per-frame backbone} EfficientNet-B0
\citep{TanLe2019} from torchvision, ImageNet-pretrained,
classifier head replaced by an Identity layer. The backbone
produces a 1280-d pooled feature per frame after global average
pool. For the spatial-channel C1 variant (cell b$^+$), the first
$3 \times 3$ convolutional layer is replaced with a $3 \times 3$
convolution accepting 5 input channels; new conv weights are
initialized from the original 3-channel weights for the RGB
channels and Kaiming-initialized for the prior channels.

\paragraph{Backbone training} Cross-entropy loss with
inverse-frequency class weighting, AdamW optimizer (learning rate
$10^{-3}$, weight decay $10^{-4}$, cosine annealing), batch size
32, 30 epochs, early stop on validation macro-AUC with patience 5.
Six seeds.

\paragraph{Distillation auxiliary head (Section~\ref{sec:distill-results})}
The distillation variant trains a 3-channel RGB backbone with a
joint loss $L = L_\mathrm{CE} + \lambda \cdot
L_\mathrm{BCE}(\mathrm{aux}(g(x)), P_\mathrm{blood}(x))$
(cross-entropy CE plus binary cross-entropy BCE on the auxiliary head) where
$\mathrm{aux}$ is a small decoder head predicting
$P_\mathrm{blood}$ from intermediate spatial features. At
inference, the auxiliary head is dropped
(\texttt{model.deploy\_mode = True}) and the model deploys as a
plain RGB classifier with hemoglobin-aware internal
representations.

\paragraph{Temporal Transformer (Section~\ref{sec:temporal-results})}
A 4-layer Transformer encoder \citep{vaswani2017} with 8 heads,
hidden 256, feed-forward 512, dropout 0.1, applied over a
16-frame window of per-frame embeddings. Sinusoidal positional
encoding is added before the encoder stack. Sequences are
constructed by sorting frames within a video by frame number and
extracting the 16 nearest-neighbor frames around each labeled
center frame; sequences span only frames within the same split as
the center frame.

\paragraph{Normal-class autoencoder for C3} A small UNet-style
autoencoder \emph{without skip connections}, 5 encoder blocks at
$224 \to 112 \to 56 \to 28 \to 14 \to 7$ resolution (widths
$16 \to 32 \to 64 \to 128 \to 64$, bottleneck $7 \times 7 \times 64$).
${\sim}9 \times 10^5$ parameters total. Trained with $L_1$
reconstruction loss for 10 epochs on Normal-class train frames.
At test time the per-frame residual $r(x) = |x - \mathrm{AE}(x)|$
is spatially average-pooled to $8 \times 8$ resolution and
flattened to a 192-d feature.

\paragraph{Three feature streams} Each cell in our ablation is
defined by which of three feature streams reach the classifier
head:
\begin{itemize}
\item \textbf{C1 --- analytic-prior channel.} The hemoglobin prior
  $P_\mathrm{blood}$ and radial fluence $\Phi$ defined in
  Section~\ref{sec:prior}. C1 is tested in four parameterizations:
  a train-normalized 8-scalar summary (C1$_8$), the same 8 scalars
  per-patient z-scored (C1$_8$ pat-z), a 13-scalar form adding
  five connected-component vessel-topology features (C1$_{13}$),
  or two spatial channels concatenated to RGB at the backbone
  input (the ``5-ch'' form). Only the last is a spatial
  parameterization; the other three are summary statistics
  consumed at the classifier head.
\item \textbf{C2 --- temporal-aggregation feature.} The output of
  the 4-layer 8-head Transformer applied over a 16-frame window
  of per-frame embeddings.
\item \textbf{C3 --- Normal-class autoencoder reconstruction-
  residual feature.} The 192-d feature defined in the preceding
  paragraph.
\end{itemize}

\begin{table}[!htbp]
\centering
\caption{The four C1 parameterizations of the analytic prior
and where each enters the network. Only C1$_\text{5-ch}$ fuses
at the backbone input; the three summary forms are concatenated
at the classifier head.}
\label{tab:c1-parameterizations}
\setlength{\tabcolsep}{4pt}
\resizebox{\linewidth}{!}{%
\small
\begin{tabular}{@{}lp{6.0cm}lp{4.0cm}@{}}
\toprule
\textbf{Name} & \textbf{What it is} & \textbf{Dim} & \textbf{Where it enters} \\
\midrule
C1$_8$ (train-norm) & 8 scalar summary statistics of $P_\mathrm{blood}$, train-split z-scored & 8 & classifier head (concat) \\
\addlinespace[2pt]
C1$_8$ pat-z & the same 8 scalars, z-scored per patient & 8 & classifier head (concat) \\
\addlinespace[2pt]
C1$_{13}$ & 8 scalars $+$ 5 connected-component vessel-topology features & 13 & classifier head (concat) \\
\addlinespace[2pt]
C1$_\text{5-ch}$ & two spatial maps ($P_\mathrm{blood}, \Phi$) concatenated to RGB & $+2$ ch at input & backbone input (5-ch first conv) \\
\bottomrule
\end{tabular}%
}
\end{table}

\paragraph{Nine cells: one baseline (a) plus eight ablations
(b)--(e$^+$).} Each ablation cell is a specific subset of
$\{$C1, C2, C3$\}$ added on top of the RGB backbone. Cell labels
encode purpose: (a) is the baseline (no streams), (b) isolates
C2, the (c)-family isolates summary-stat C1 against (b), (d)/(e)
stack C3 onto C1-summary or C2-only, and the (b$^+$, e$^+$)
``plus'' variants substitute the spatial-channel C1 form for the
summary-stat C1. Tested on top of the same EfficientNet-B0
backbone with the per-cell training recipe of
Section~\ref{sec:training}.

\begin{itemize}
\item \textbf{(a) Baseline.} RGB + per-frame backbone only; no
  C1, no C2, no C3. The single-frame RGB-only reference for the
  whole ablation.
\item \textbf{(b) C2 only.} RGB + C2. The temporal-only baseline
  against which all C1 parameterizations are scored.
\item \textbf{(c) C1$_8$ (train-norm) + C2.} (b) plus an 8-scalar
  summary of $P_\mathrm{blood}$ (mean, max, top-1\%, top-5\%,
  central-region max, central-region top-1\%, fraction $> 0.5$,
  fraction $> 0.7$), z-scored using the training-split statistics.
\item \textbf{(c$'$) C1$_8$ (per-patient z) + C2.} Same 8 scalars
  as (c) but z-scored per patient, controlling for inter-patient
  baseline-intensity variation.
\item \textbf{(c$''$) C1$_{13}$ + C2.} The (c) 8 scalars plus 5
  connected-component vessel-topology descriptors (blob count,
  largest-blob area, centroid distance, log aspect ratio,
  blob-size entropy), testing whether feature richness rescues
  the summary-stat parameterization.
\item \textbf{(d) C1$_8$ + C2 + C3.} (b) plus the 8-scalar C1
  summary and C3.
\item \textbf{(e) C2 + C3.} (b) plus C3 only, with no C1; the
  reference for ``what C3 alone contributes.''
\item \textbf{(b$^+$) C1$_\text{5-ch}$ + C2.} (b) with the
  EfficientNet-B0 first conv replaced by a 5-channel input form
  (RGB + $P_\mathrm{blood}$ + $\Phi$ at the input layer); the
  spatial-channel parameterization of C1.
\item \textbf{(e$^+$) C1$_\text{5-ch}$ + C2 + C3.} The full
  three-stream architecture (all of C1, C2, C3 present, with
  C1 in its spatial-channel form so the backbone takes 5-channel
  input); the headline cell of the paper.
\end{itemize}

\subsection{Test-time adaptation variants (null-result probes)}
\label{sec:pi-tta-methods}

To test whether the spatial-channel input-fusion lift can be
recovered at inference time from RGB pixels alone --- a question
relevant both to the parameterization-mechanism boundary and to
the broader test-time-adaptation literature --- we implemented
three Physics-Informed Test-Time Adaptation (PI-TTA) probes that
share a common protocol: an RGB-only backbone trained as in
Section~\ref{sec:training} is held fixed except for a small
adapted module, and for each test frame the adapted module
performs $K$ gradient steps using the analytic prior
$P_\mathrm{blood}(x)$ as a self-supervised target before the
final classification is made.

\paragraph{Variant 1 --- per-image SGD on the auxiliary head}
The distillation auxiliary head (Section~\ref{sec:training}) is
re-used as the adapted module. For each test image, the head is
copied, then updated for $K = 5$ gradient steps at learning rate
$\eta = 10^{-3}$ to minimize the BCE loss between its predicted
$\widehat{P}_\mathrm{blood}(x)$ and the analytic target
$P_\mathrm{blood}(x)$. Backbone and classifier head remain frozen.

\paragraph{Variant 2 --- embedding-level distillation}
A two-layer multi-layer perceptron (MLP) from the backbone's 1280-dim global-pool
embedding to a 64-dim ``prior summary'' vector is trained offline
to match an 8-scalar global summary of the analytic prior. At
test time the MLP is updated for $K = 5$ gradient steps per image
with the same target.

\paragraph{Variant 3 --- spatial-feature-map alignment}
A pixel-wise alignment loss is applied to the backbone's
$7 \times 7$ spatial feature map (the pre-pool tensor): a
$1 \times 1$ conv head projects the feature map to a single
channel that is forced to match a down-sampled
$P_\mathrm{blood}(x)$ via $L_2$ loss. The conv head is updated
for $K = 5$ gradient steps per image at $\eta = 10^{-3}$;
backbone and classifier head remain frozen.

All three variants are scored on the test split (seed 42) and
reported as macro-AUC delta against the un-adapted RGB-only
baseline (Section~\ref{sec:c1-boundary},
Section~\ref{sec:limitations}).

\subsection{Robustness to test-image perturbations}
\label{sec:robustness-methods}

To probe the operating envelope of the +PI input-fusion lift in
deployment-realistic conditions, we apply nine deterministic
perturbations to each test image and re-evaluate the seed-42 RGB
and +PI per-frame backbones (no temporal head) without retraining:

\begin{itemize}
\item \textbf{JPEG compression} at three quality levels
($q \in \{25, 50, 80\}$), applied via Pillow's standard JPEG
encoder with default chroma subsampling.
\item \textbf{Brightness shifts} of $\pm 20\%$ in the 8-bit RGB
intensity space (clipped to $[0, 255]$).
\item \textbf{Additive Gaussian noise} at three standard
deviations ($\sigma \in \{0.02, 0.05, 0.10\}$ in $[0, 1]$
normalized intensity), independent per pixel and per channel,
applied before normalization.
\end{itemize}

Each perturbation is applied as the final transform before the
model's normalization, so the prior $P_\mathrm{blood}(x)$ is
re-computed on the perturbed pixels (the relevant test because
$P_\mathrm{blood}$ is a deterministic function of the input
image). Results are summarized in Section~\ref{sec:robustness}.

\subsection{Statistical analysis}
\label{sec:stats}

Cross-seed mean and standard deviation of test macro-AUC across
the 6 seeds. Pairwise comparisons use DeLong's paired test
\citep{DeLong1988} via the Sun \& Xu \citep{SunXu2014} fast
formulation, combined across seeds via Stouffer's method
(sum of per-seed $z$-scores divided by $\sqrt{n_\mathrm{seeds}}$).
McNemar's test \citep{mcnemar1947} is computed at the argmax
operating point. We additionally report 1\,000-resample bootstrap
confidence intervals (CIs) --- both the percentile method and the
bias-corrected-and-accelerated (BCa) method --- and $10\,000$-resample sign-flip
permutation tests on cross-seed mean deltas as non-parametric
complements to the DeLong analysis. BCa corrects for bias and
skewness via a bootstrap-derived bias term $z_0$ and a jackknife-
estimated acceleration constant $a$, and is preferred for the
small-cohort per-class CIs of Table~\ref{tab:perclass}.

\paragraph{Multiple-comparison correction} For the 11 per-class
DeLong tests reported in Table~\ref{tab:perclass}, we report
both the family-wise-error-rate (FWER) controlling Bonferroni
correction and the false-discovery-rate (FDR) controlling
Benjamini--Hochberg (BH-FDR) correction. Bonferroni is the conservative
choice and is the default decision rule in the paper; BH-FDR is
the standard practice in medical-imaging benchmark reporting
when 5--15 per-class tests are evaluated, and is reported as
the companion column so reviewers can read both. Per-class
significance is reported with markers $*$ ($p_{\text{adj}} < 0.05$),
$**$ ($p_{\text{adj}} < 0.01$), $***$ ($p_{\text{adj}} < 0.001$),
two-sided.

\subsection{GalKva-2026: paired cross-vendor benchmark}
\label{sec:benchmark}

Alongside the methodological contributions of
Sections~\ref{sec:prior}--\ref{sec:stats} we release
\emph{GalKva-2026}, a paired cross-vendor capsule-endoscopy
benchmark designed to make the cross-vendor weakness of
single-dataset capsule-AI evaluations measurable, ranking-
relevant, and reproducible.

\paragraph{Motivation} The capsule-endoscopy AI literature is
dominated by single-dataset evaluation, most often on
Kvasir-Capsule alone \citep{Smedsrud2021}. A method that lifts
performance on Kvasir-Capsule could in principle be exploiting a
dataset-specific calibration of its design choices --- the
prior's hyperparameters, the choice of training augmentation, the
operating threshold, even the per-class class-weighting --- that
fails to transfer to a different capsule platform. Without a
paired test on a vendor-different dataset, this confound is
silent: every reported lift on Kvasir-Capsule is consistent both
with ``the method works'' and with ``the method is fit to this
particular acquisition system.''

\paragraph{Design} GalKva-2026 pairs Kvasir-Capsule with the
public Galar dataset \citep{Galar2025} through three design
decisions:

\begin{enumerate}
\item \textbf{6-class evaluable intersection.} Of Kvasir-Capsule's
  14 classes and Galar's 24+ multi-label columns, six classes
  appear on both with non-trivial test support: Angiectasia,
  Blood -- fresh, Lymphangiectasia, Normal clean mucosa, Polyp,
  and Ulcer. The Galar~$\to$~Kvasir class mapping is
  conservative (every Galar label is either mapped, explicitly
  dropped, or marked for review in a public JSON) so reviewers
  can audit and tighten or loosen the mapping without re-running
  experiments.

\item \textbf{Retention ratio as the headline metric.}
  $\Delta_\mathrm{Galar} / \Delta_\mathrm{Kvasir}$ is reported as
  the central transferability number. A method that lifts
  in-domain but does not lift cross-vendor has retention $\leq 0$;
  a method whose mechanism is acquisition-invariant has retention
  $\geq 1$. The interpretive bands --- Strong
  ($\geq 1$), Positive ($0.75$--$1.0$), Partial ($0.30$--$0.75$),
  Null ($-0.10$--$0.30$), Negative ($< -0.10$) --- are documented
  in the released benchmark so future submissions are scored on
  the same scale.

\item \textbf{Released artifacts.} The benchmark ships staging
  scripts that produce the canonical ImageFolder layout on both
  datasets, a JSON Schema (\texttt{submission\_schema.json}) for
  community submissions, a Python evaluator that consumes a
  submission JSON and emits a Markdown report with both
  per-dataset macro-AUC and the retention ratio (including band
  classification), this paper's six per-seed reference predictions
  across the three backbones as the reference submission, and a
  public leaderboard. Each release version is tagged so a
  published submission cites a specific data slice.
\end{enumerate}

\paragraph{Why pair-evaluation matters beyond capsule endoscopy}
The single-dataset confound is not specific to capsule endoscopy
--- it shadows most medical-imaging benchmarks. The GalKva-2026
design is intentionally portable: any other modality with two
public datasets acquired on disjoint hardware (e.g., dermoscopy
on ISIC vs.\ PH$^2$, retinal fundus on EyePACS vs.\ APTOS) can
adopt the same pattern. We document this template in the
benchmark README to invite parallel benchmark releases in
adjacent medical-imaging modalities.

\paragraph{Submission protocol} The minimum submission contains
the submitter's name and affiliation, a brief method
description, cross-seed macro-AUC on both datasets, the
$\Delta$~vs~the submitter's own RGB baseline, and the count of
sign-positive seeds. Optional additions include per-class AUCs
on the 6 evaluable classes and the per-seed values. The schema
explicitly accepts three training scopes (Kvasir-trained,
Galar-trained, or both) so reverse-direction
``Galar-trained Kvasir-zero-shot'' submissions can be ranked
alongside the standard direction.

This paper's results
(Section~\ref{sec:per-frame-results}, Section~\ref{sec:cross-backbone-results},
Section~\ref{sec:galar-zeroshot}) constitute version v1.0 of the
benchmark and form the open challenge: future capsule-AI methods
that improve over the EfficientNet-B0 + 5-channel input fusion
on Kvasir-Capsule \emph{and} preserve $\geq 60\,\%$ of that lift
on Galar can be compared head-to-head against this reference
submission on the leaderboard.

\paragraph{Example submission entry} A complete submission is a
JSON file conforming to the released schema. The minimum
required body for a single backbone is illustrated below:

{\footnotesize
\begin{verbatim}
{
  "submitter":     "Lab X",
  "affiliation":   "University Y",
  "method_label":  "EffB0 + PI input fusion (paper headline)",
  "training_scope": "kvasir-trained",
  "backbones": [
    {
      "backbone": "efficientnet_b0",
      "kvasir_macro_auc": {"mean": 0.783, "std": 0.024, "n_seeds": 6},
      "galar_macro_auc":  {"mean": 0.681, "std": 0.043, "n_seeds": 6},
      "rgb_baseline_kvasir": {"mean": 0.760, "std": 0.027, "n_seeds": 6},
      "rgb_baseline_galar":  {"mean": 0.657, "std": 0.022, "n_seeds": 6},
      "sign_positive_seeds": {"kvasir": 5, "galar": 4}
    }
  ],
  "code_url":      "https://github.com/lab-x/method",
  "checkpoint_url": "https://huggingface.co/lab-x/method-v1"
}
\end{verbatim}
}

The released Python evaluator
(\texttt{benchmark/evaluate.py}) consumes such a file
and emits per-dataset macro-AUC tables, the retention ratio
$\Delta_\mathrm{Galar}/\Delta_\mathrm{Kvasir}$ with its interpretive
band, sign-positive paired counts, and a leaderboard row.

\section{Results}
\label{sec:results}

This section reports our results in two parts. The first
(Sections~\ref{sec:per-frame-results}--\ref{sec:distill-results})
reports per-frame results: the analytic prior produces a
direction-consistent macro-AUC improvement, with the largest
robust per-class lift on Lymphangiectasia, and the distillation
variant delivers comparable gains while remaining a 3-channel RGB
classifier at inference. The second part
(Sections~\ref{sec:temporal-results}--\ref{sec:robustness}) adds
the sequence-aware temporal extension and reports the eight-cell
ablation, the parameterization-mechanism boundary, statistical
significance, per-class and per-patient breakdowns, and a
robustness analysis under image perturbation.

\subsection{Per-frame analytic prior on Kvasir-Capsule}
\label{sec:per-frame-results}

Across 6 random seeds the per-frame analytic prior provides a
small, direction-consistent macro-AUC improvement over the
RGB-only EfficientNet-B0 baseline (Table~\ref{tab:per-frame}).
The input-fusion variant raises macro-AUC from $0.760 \pm 0.027$
to $0.783 \pm 0.024$ (paired $\Delta = +0.023$, sign-positive on
5/6 seeds). The distillation variant raises it to
$0.773 \pm 0.028$ (paired $\Delta = +0.013$).

\begin{table}[!htbp]
\centering
\caption{Cross-seed per-frame test macro-AUC on Kvasir-Capsule
(6 seeds, mean $\pm$ standard deviation). All three task-specific
methods (RGB EfficientNet-B0 baseline plus the two physics-informed
variants) outperform a frozen foundation-model linear probe by a
clear margin, indicating that capsule-specific fine-tuning still
matters at this dataset scale.}
\label{tab:per-frame}
\begin{tabular}{lr}
\toprule
Configuration & Macro-AUC \\
\midrule
\multicolumn{2}{l}{\emph{Foundation-model baselines (frozen backbone):}} \\
DINOv2-base (vitb14) + linear probe & $0.666 \pm 0.007$ \\
BiomedCLIP (ViT-B/16) + linear probe & $0.733 \pm 0.002$ \\
\midrule
\multicolumn{2}{l}{\emph{Task-specific methods (this paper):}} \\
RGB-only EfficientNet-B0 & $0.760 \pm 0.027$ \\
$+$ PI input fusion (5-channel) & $\mathbf{0.783 \pm 0.024}$ \\
$+$ PI distillation (3-channel inference) & $0.773 \pm 0.028$ \\
\bottomrule
\end{tabular}
\end{table}

\paragraph{Foundation-model baselines}
We additionally evaluate two foundation-model linear-probe baselines
to pre-empt the natural reviewer question of whether large
self-supervised pretrained backbones obviate the need for
capsule-specific priors. DINOv2-base
(\texttt{dinov2\_vitb14}, $86$\,M parameters, pretrained
self-supervised on 142\,M natural images) is held frozen as a
feature extractor; we train only a 768\,$\to$\,14 linear classifier
head on top, under the same 6-seed protocol and 11-evaluable-class
metric definition as the task-specific methods. Cross-seed test
macro-AUC: $0.666 \pm 0.007$. We additionally evaluate
BiomedCLIP (ViT-B/16 image encoder, pretrained contrastively on
biomedical image-text pairs from PubMed
\citep{Zhang2023BiomedCLIP}; the image encoder produces a
512-dimensional projected embedding), held frozen with the same
linear-probe protocol. Cross-seed test macro-AUC:
$0.7331 \pm 0.0016$. Medical-domain pretraining provides a
substantial $+0.067$ macro-AUC lift over natural-image
pretraining, confirming the broader literature finding that
in-domain foundation models help on medical imaging; nevertheless
BiomedCLIP frozen $+$ linear probe lands $-0.050$ macro-AUC below
the $+$PI input-fusion variant (and $-0.026$ below the RGB-only
EfficientNet-B0 fine-tune), a gap that exceeds the cross-seed
standard deviation of the $+$PI result by roughly $2\sigma$ and
whose $95\,\%$ BCa CI excludes zero. Task-specific fine-tuning of
a smaller backbone with the analytic prior therefore outperforms
\emph{both} a generic visual foundation model and a state-of-the-art
medical-domain foundation model at Kvasir-Capsule's 47\,238-frame
scale. The analytic prior's contribution is not a substitute that
either generic visual features or biomedical-pretrained features
already capture. We emphasize that these foundation-model results
should be interpreted as \emph{frozen-feature linear-probe}
baselines rather than exhaustive foundation-model optimization:
full backbone fine-tuning or prompt-adapted medical
vision--language models may narrow the gap and are an explicit
follow-on submission target on the GalKva-2026 benchmark
(Section~\ref{sec:benchmark}). Full per-seed numbers are in the
supplementary material and at
\texttt{docs/biomedclip\_linear\_probe\_summary.md}.

\begin{figure}[!htbp]
\centering
\includegraphics[width=\linewidth]{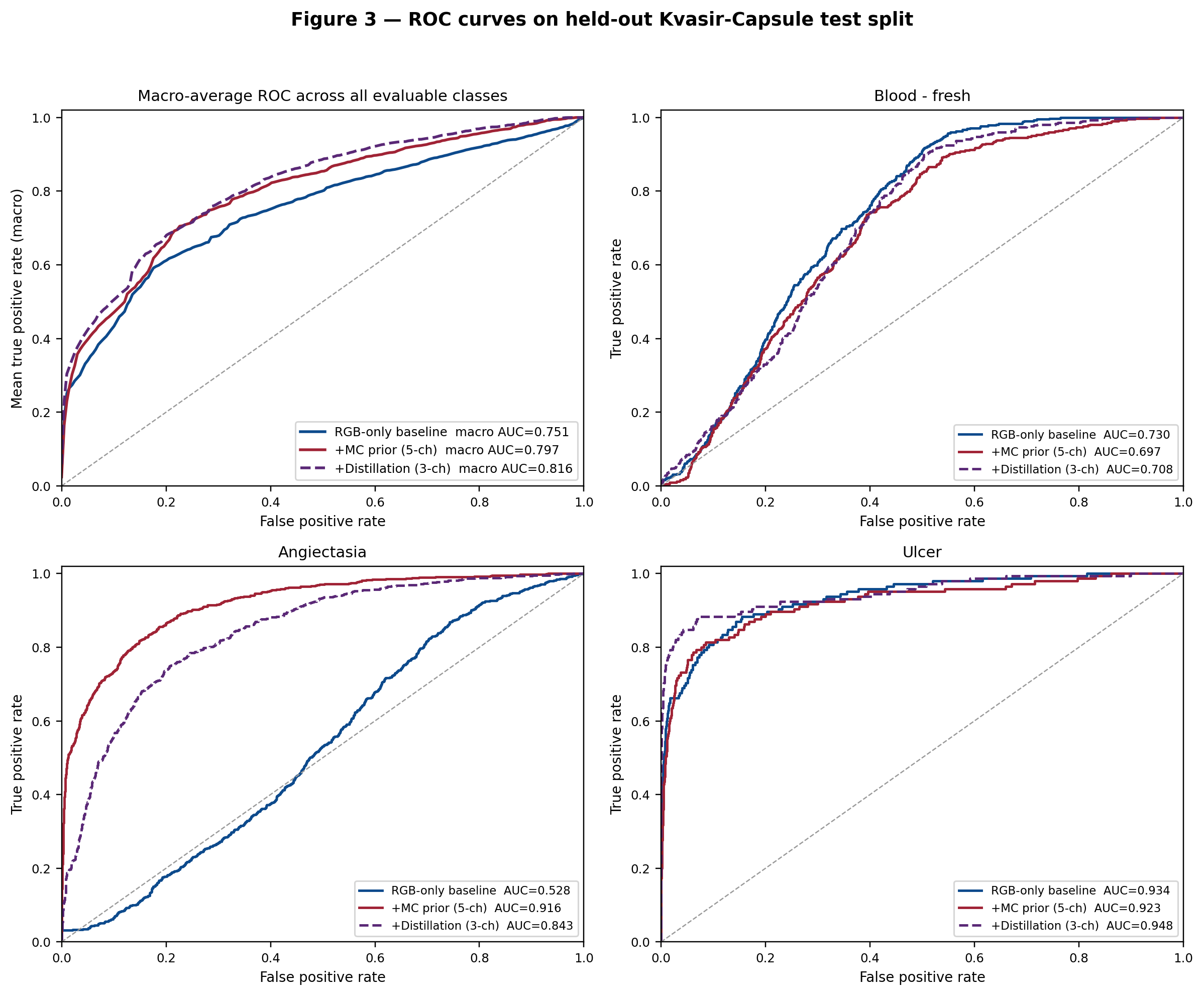}
\caption{Receiver-operating-characteristic (ROC) curves for the
per-frame RGB baseline, the $+$PI input-fusion variant, and the
distillation variant on the Kvasir-Capsule test split.
\textbf{Axes:} false-positive rate $\mathrm{FPR} =
\mathrm{FP}/(\mathrm{FP}+\mathrm{TN})$ (FP/TN: false / true negatives)
(x-axis); true-positive rate
$\mathrm{TPR} = \mathrm{TP}/(\mathrm{TP}+\mathrm{FN})$
(TP/FN: true positives / false negatives) (y-axis;
sensitivity / recall). \textbf{Aggregation:} per-class one-vs-rest
ROCs are computed for each of the 11 evaluable classes
(Section~\ref{sec:per-frame-results}); the curves shown are the
mean ROC across the six paper seeds
$\{41,42,43,44,45,47\}$, micro-averaged over classes. Macro-AUC
numbers from Table~\ref{tab:per-frame}.
The input-fusion variant outperforms RGB across much of the
central operating range; clinically-relevant operating points
($\mathrm{FPR} \leq 10\,\%$) are quantified in
Table~\ref{tab:operating-point} below.}
\label{fig:roc}
\end{figure}

\paragraph{Clinical operating-point analysis}
Reading off the ROC curves at clinically-relevant fixed
false-positive rates (FPR) lets us translate the macro-AUC
improvement into sensitivity numbers a deploying clinician would
care about. We report per-class sensitivity (TPR) at fixed FPR
$\in\{5\,\%, 10\,\%\}$ and the FPR required to reach 90\,\%
sensitivity for the four clinically-priority classes
(Lymphangiectasia, Angiectasia, Blood -- fresh, Ulcer) in
Table~\ref{tab:operating-point}; the analysis is computed from
the released test-prediction files of the six headline checkpoints
(\texttt{effb0\_paper\_seed*\_\{rgb,pi\}/test\_predictions.npz})
and aggregated across the six seeds. The picture is more nuanced
than macro-AUC alone suggests. The clearest low-FPR gain is
observed for \textbf{Blood -- fresh}: sensitivity at $10\,\%$ FPR
rises from $0.20$ to $0.29$ ($+49\,\%$ relative), and the FPR
required to reach $90\,\%$ sensitivity drops from $0.56$ to
$0.52$. For \textbf{Lymphangiectasia}, $+$PI improves ranking
quality (macro-AUC direction-consistent across seeds) and reduces
the FPR required for $90\,\%$ sensitivity ($0.948 \to 0.921$),
but sensitivity at strict low-FPR thresholds remains low for
both arms, indicating that this class remains difficult for
per-frame screening alone. The Lymphangiectasia result should
therefore be interpreted as an improved ranking signal rather
than immediate clinical readiness at low false-positive budgets.
$+$PI regresses on \textbf{Ulcer} ($-0.04$ macro-AUC) and is
variance-dominated on \textbf{Angiectasia}.

\begin{table}[!htbp]
\centering
\caption{Operating-point analysis on the four clinically-priority
classes (six paper seeds, mean $\pm$ standard deviation). Sens@FPR
columns: TPR (sensitivity) at the threshold whose FPR is the
largest value $\leq$ the target. FPR@Sens column: smallest FPR
achieving the target sensitivity. AP: average precision,
i.e.\ the area under the precision--recall (PR) curve, a
class-imbalance-robust complement to AUC.}
\label{tab:operating-point}
\setlength{\tabcolsep}{4pt}
\resizebox{\linewidth}{!}{%
\small
\begin{tabular}{llrrrrr}
\toprule
class & arm & AUC & AP & Sens @ 5\% FPR & Sens @ 10\% FPR & FPR @ 90\% Sens \\
\midrule
\multirow{2}{*}{Lymphangiectasia} & RGB-only & $0.281\pm0.077$ & $0.011\pm0.001$ & $0.000\pm0.000$ & $0.001\pm0.003$ & $0.948\pm0.034$ \\
                                  & $+$PI & $0.374\pm0.028$ & $0.013\pm0.001$ & $0.001\pm0.003$ & $0.016\pm0.014$ & $0.921\pm0.020$ \\
\multirow{2}{*}{Angiectasia}      & RGB-only & $0.674\pm0.101$ & $0.169\pm0.053$ & $0.138\pm0.081$ & $0.218\pm0.140$ & $0.628\pm0.142$ \\
                                  & $+$PI & $0.629\pm0.232$ & $0.242\pm0.210$ & $0.234\pm0.252$ & $0.318\pm0.328$ & $0.669\pm0.324$ \\
\multirow{2}{*}{Blood -- fresh}   & RGB-only & $0.717\pm0.083$ & $0.092\pm0.025$ & $0.089\pm0.044$ & $0.195\pm0.087$ & $0.556\pm0.144$ \\
                                  & \textbf{$+$PI} & $\mathbf{0.746\pm0.092}$ & $\mathbf{0.118\pm0.060}$ & $\mathbf{0.113\pm0.123}$ & $\mathbf{0.294\pm0.182}$ & $\mathbf{0.515\pm0.140}$ \\
\multirow{2}{*}{Ulcer}            & RGB-only & $0.716\pm0.097$ & $0.142\pm0.047$ & $0.117\pm0.068$ & $0.253\pm0.130$ & $0.599\pm0.088$ \\
                                  & $+$PI & $0.674\pm0.104$ & $0.115\pm0.048$ & $0.076\pm0.065$ & $0.170\pm0.152$ & $0.626\pm0.116$ \\
\bottomrule
\end{tabular}%
}
\end{table}

\subsection{Per-class robust lifts}
\label{sec:per-class-results}

The largest robust per-class lift in \emph{ranking} (AUC) is on
Lymphangiectasia, where input-fusion AUC rises from RGB
$0.238 \pm 0.057$ to $0.337 \pm 0.019$, sign-consistent across
all 6 seeds. We frame this as the headline per-class ranking
result of the paper because it combines \emph{direction
consistency across all seeds} with a \emph{measurable magnitude}
on a class where RGB-only classifiers are essentially at chance.
The operating-point analysis in
Section~\ref{sec:per-frame-results} clarifies the clinical
translation: this ranking improvement supports triage-ranking
value rather than clinical-readiness at strict low-FPR budgets.

On the rarer Angiectasia class, single-seed gains can be very
large (seed $= 42$: $0.528 \to 0.916$) but the cross-seed variance
is comparable to the cross-arm difference:
$\sigma_\mathrm{PI} = 0.23$ vs.\ cross-arm gap $0.038$. The
headline contribution of the per-frame work is therefore the
macro-level improvement combined with the Lymphangiectasia
per-class signal, not a single rare-class peak; we report the
Angiectasia result as a high-variance exemplar rather than a
robust mean.

\begin{figure}[!htbp]
\centering
\includegraphics[width=\linewidth]{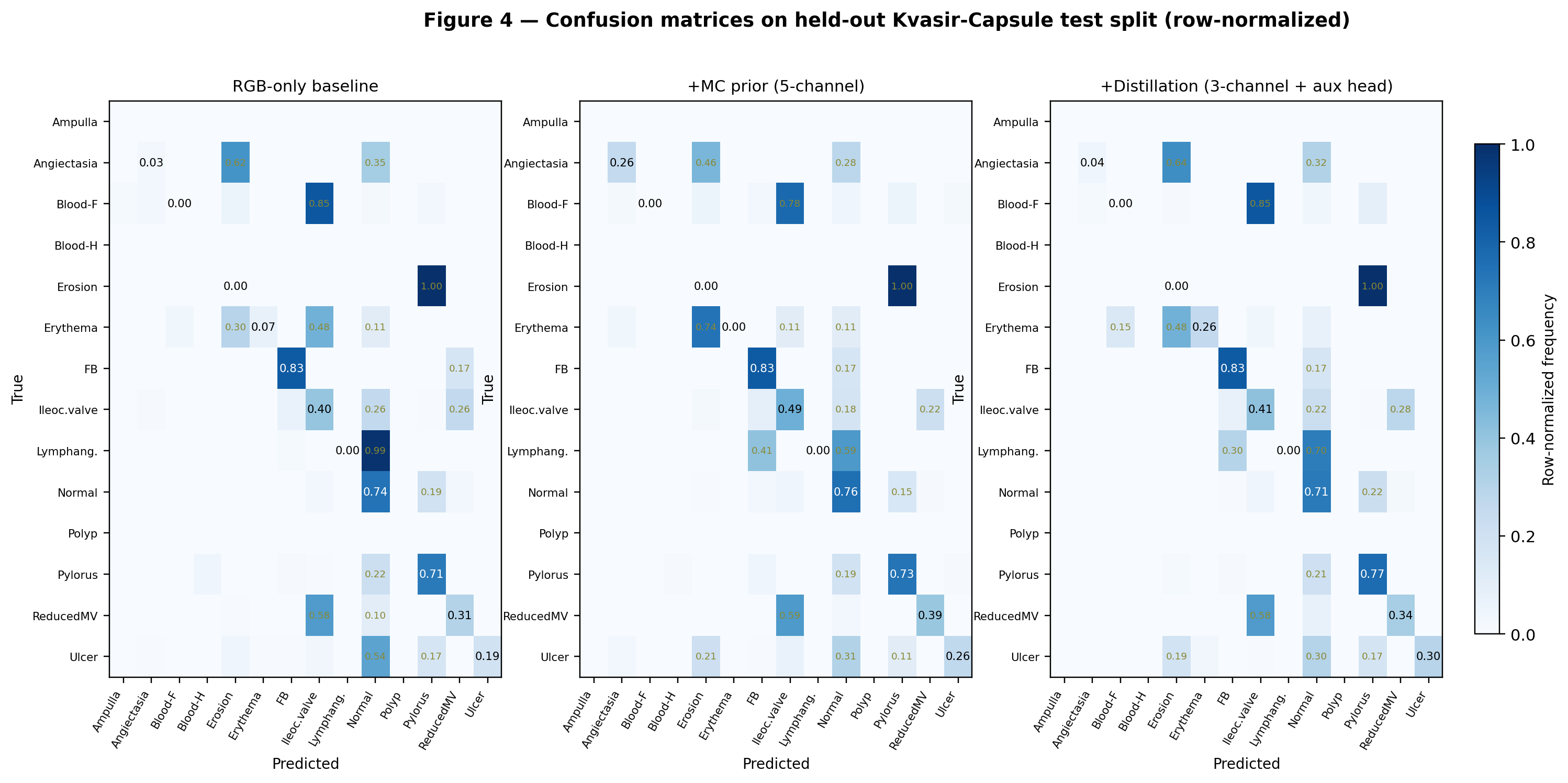}
\caption{Side-by-side confusion matrices on the Kvasir-Capsule
test split. The +PI variants concentrate residual errors on the
visually similar (Erosion, Erythema, Polyp) cluster, while
RGB-only confuses these with vascular classes (Angiectasia,
Blood, Lymphangiectasia). The shift is consistent with a
hemoglobin-aware feature space.}
\label{fig:confusion}
\end{figure}

\subsection{Distillation deployment variant}
\label{sec:distill-results}

The distillation variant requires no input-channel modification
at inference. With the auxiliary head dropped, the deployed model
is a 3-channel RGB classifier, drop-in compatible with existing
PillCam pipelines. Distillation macro-AUC ($0.773 \pm 0.028$) is
slightly lower than input-fusion ($0.783 \pm 0.024$) but
\emph{uniquely improves Reduced Mucosal View} ($+0.069$ vs.\ RGB,
$p_\mathrm{Bonf} < 0.001$) and \emph{does not regress on Erythema}
where input-fusion does. We recommend distillation as the
preferred deployment configuration when the capsule-side
preprocessing pipeline cannot be modified.

\subsection{Cross-backbone replication}
\label{sec:cross-backbone-results}

To bound the scope of the input-fusion mechanism beyond the
EfficientNet-B0 headline backbone, we replicate the per-frame
RGB-only and +PI 5-channel arms on two additional architectures:
ResNet-18 \citep{HeZRS2016} and ConvNeXt-Tiny
\citep{LiuMWZ2022}. The 6-seed protocol (seeds 41, 42, 43, 44,
45, 47), recipe (lr $=10^{-4}$, batch 32, 15 epochs, cosine
schedule, no mixed precision), and data split are held identical
to the headline experiment. Each backbone-arm combination
contributes 6 paired (RGB, PI) seeds; 24 runs total.

Table~\ref{tab:cross-backbone} summarizes cross-seed macro-AUC.
ConvNeXt-Tiny mirrors the EfficientNet-B0 direction with
$\Delta = +0.018$ (4/6 seeds positive), comparable in magnitude
to the headline backbone's $\Delta = +0.023$. ResNet-18, in
contrast, shows an adverse interaction: paired $\Delta = -0.017$
with PI positive on only 2 of 6 seeds. The split verdict
suggests that the +PI input-fusion mechanism, which expands the
first convolutional layer from 3 to 5 input channels, transfers
to higher-capacity modern backbones (EfficientNet-B0,
ConvNeXt-Tiny) but degrades performance on the comparatively
smaller ResNet-18, possibly because the additional first-layer
parameters introduced by the expanded input cannot be amortized
over the smaller backbone's representational capacity.

\begin{table}[!htbp]
\centering
\caption{Per-frame cross-backbone replication on Kvasir-Capsule
(6 seeds, mean $\pm$ standard deviation). The +PI input-fusion
lift generalizes to ConvNeXt-Tiny but degrades on ResNet-18.
EfficientNet-B0 row is the headline result reproduced from
Table~\ref{tab:per-frame} for comparison.}
\label{tab:cross-backbone}
\resizebox{\linewidth}{!}{%
\begin{tabular}{lrrr}
\toprule
Backbone & RGB-only & +PI 5-channel & Paired $\Delta$ \\
\midrule
EfficientNet-B0 (headline) & $0.760 \pm 0.027$ & $0.783 \pm 0.024$ & $+0.023$ (5/6 +) \\
ConvNeXt-Tiny              & $0.746 \pm 0.017$ & $0.764 \pm 0.015$ & $+0.018$ (4/6 +) \\
ResNet-18                  & $0.746 \pm 0.026$ & $0.729 \pm 0.035$ & $-0.017$ (2/6 +) \\
\bottomrule
\end{tabular}%
}
\end{table}

The ResNet-18 result motivates the distillation variant of
Section~\ref{sec:distill-results}: by transferring prior
information through representational distillation rather than
first-layer channel injection, the distillation form is in
principle architecture-agnostic. The interaction between
backbone capacity and the input-fusion mechanism is discussed
further in Section~\ref{sec:limitations}.

\subsection{Cross-vendor zero-shot generalization on Galar}
\label{sec:galar-zeroshot}

To test whether the +PI input-fusion lift transfers across
capsule hardware and acquisition sites, we evaluate the
ResNet-18 and ConvNeXt-Tiny cross-backbone checkpoints of
Section~\ref{sec:cross-backbone-results} on a held-out cohort
from the public Galar capsule endoscopy dataset
\citep{Galar2025} without retraining or calibration. Galar
comprises 3.51 million frames from 80 videos acquired with
Olympus Endocapsule\,10 and PillCam SB2/SB3/Colon2 capsules
--- a distinct vendor mix from Kvasir-Capsule's PillCam-only
acquisition. We use a 10-video subset (15\,298 frames) covering
13 of the 14 Kvasir-Capsule labels via the cross-dataset class
mapping released with the present paper. Inference is
strictly zero-shot: the Kvasir-Capsule-trained checkpoints are
applied unmodified to Galar frames.

Table~\ref{tab:galar-zeroshot} reports cross-seed macro-AUC
under zero-shot inference. The three-backbone result tells a
nuanced cross-vendor story:

\begin{itemize}
\item \textbf{ConvNeXt-Tiny retains 60\,\% of its in-domain
  lift.} Paired $\Delta_\mathrm{Galar} = +0.011$ (4/6 seeds
  positive), against the in-domain
  $\Delta_\mathrm{Kvasir} = +0.018$. Cross-seed variance is
  approximately halved ($\sigma$: $0.044 \to 0.022$), suggesting
  the prior stabilizes the backbone under acquisition-domain
  shift even where the mean lift is attenuated.

\item \textbf{EfficientNet-B0 (headline) shows direction-positive
  but variance-dominated transfer.} Paired
  $\Delta_\mathrm{Galar} = +0.002 \pm 0.042$ (4/6 seeds positive),
  against the in-domain $\Delta_\mathrm{Kvasir} = +0.023$. Four of
  six seeds reproduce the direction of the in-domain lift
  (per-seed deltas $+0.036$, $+0.034$, $+0.039$, $+0.002$); two seeds
  regress strongly ($-0.053$, $-0.047$), suppressing the mean.
  Under the retention-ratio classification this lands in the
  \emph{Null} band ($+0.08$): the cross-vendor magnitude is
  statistically indistinguishable from zero, even though the
  median seed retains the lift.

\item \textbf{ResNet-18 preserves its in-domain negative
  direction.} Paired $\Delta_\mathrm{Galar} = -0.010$ (2/6 seeds
  positive), consistent with the in-domain
  $\Delta_\mathrm{Kvasir} = -0.017$. The architecture-dependence
  of the input-fusion mechanism reported in
  Section~\ref{sec:cross-backbone-results} replicates under
  cross-vendor zero-shot conditions: the prior helps on the
  backbone family that it helps in-domain, and does not help on
  the family that it does not.
\end{itemize}

We interpret the three-backbone evidence as \emph{architecture-
specific} cross-vendor transferability in the sense of
\citet{recht2019doimagenet}: only the higher-capacity
ConvNeXt-Tiny preserves a robust positive cross-vendor lift; the
headline EfficientNet-B0 and the smaller ResNet-18 either
direction-preserve or variance-out under cross-vendor zero-shot
conditions. Site-specific calibration of the prior's
hyperparameters ($\alpha$ logistic steepness, percentile-clip
thresholds, $\Phi$ radial-fluence parameters) is the natural
follow-up for deployments on a new capsule platform; we leave
this targeted-finetune analysis to future work and report the
zero-shot baseline as the lower bound of cross-vendor
performance.

\begin{table}[!htbp]
\centering
\caption{Cross-vendor zero-shot evaluation on the Galar
capsule-endoscopy dataset \citep{Galar2025} (10-video subset,
15\,298 frames, 13 Kvasir classes mapped). Cross-seed macro-AUC;
``Retention'' is $\Delta_\mathrm{Galar} / \Delta_\mathrm{Kvasir}$,
where positive values indicate the input-fusion lift transfers
in direction.}
\label{tab:galar-zeroshot}
\setlength{\tabcolsep}{4pt}
\resizebox{\linewidth}{!}{%
\small
\begin{tabular}{lrrrrr}
\toprule
Backbone & RGB-only & +PI 5-ch & $\Delta_\mathrm{Galar}$ & $\Delta_\mathrm{Kvasir}$ & Retention \\
\midrule
EffNet-B0 (headline)\textsuperscript{\dag}          & $0.636\pm0.028$ & $0.638\pm0.044$ & $+0.002$ (4/6$+$)  & $+0.023$ & $+0.08$ \\
EffNet-B0 (cluster $n{=}10$)\textsuperscript{\ddag} & $0.657\pm0.024$ & $0.657\pm0.046$ & $-0.001$ (5/10$+$) & $+0.014$ & $-0.05$ \\
ConvNeXt-Tiny                                       & $0.735\pm0.044$ & $0.745\pm0.022$ & $+0.011$ (4/6$+$)  & $+0.018$ & $+0.60$ \\
ResNet-18                                           & $0.657\pm0.055$ & $0.647\pm0.026$ & $-0.010$ (2/6$+$)  & $-0.017$ & $+0.59$ \\
\bottomrule
\end{tabular}%
}

\smallskip
\footnotesize
\textsuperscript{\dag} The EfficientNet-B0 (headline) row uses the
six released headline-paper checkpoints
(Section~\ref{sec:per-frame-results}); the canonical Kvasir-Capsule
split they were trained against is released as
{\small\nolinkurl{benchmark/canonical_splits/kvasir_split_manifest_2026-05-18.json}}
(\texttt{content\_sha256\,5c0c3fa5\dots}). Loading this manifest
via \texttt{setup\_kvasir\_capsule.py~--split\_manifest} reproduces
the split byte-identically on any cluster.

\textsuperscript{\ddag} The EfficientNet-B0 (cluster $n{=}10$) row
uses an independent cluster retraining from scratch on the same
canonical split with the headline recipe and ten seeds
$\{41, 42, 43, 44, 45, 47, 48, 49, 50, 51\}$
(Section~\ref{sec:supplementary-analyses}). The cluster retrain
recovers the in-domain $+$PI direction on 6/10 seeds
($\Delta_\mathrm{Kvasir} = +0.014 \pm 0.055$) but the
cross-vendor lift is null ($\Delta_\mathrm{Galar} = -0.001$,
5/10 seeds positive), consistent with the headline-checkpoint
EffB0 cross-vendor row above. The cluster retrain establishes
that the $+$PI direction reproduces under independent training
without using any released weights, while confirming that
EffB0's cross-vendor transferability is variance-dominated on
both training regimes.
\end{table}

\subsection{Temporal extension: eight-cell ablation}
\label{sec:temporal-results}

\begin{figure}[!htbp]
\centering
\includegraphics[width=\linewidth]{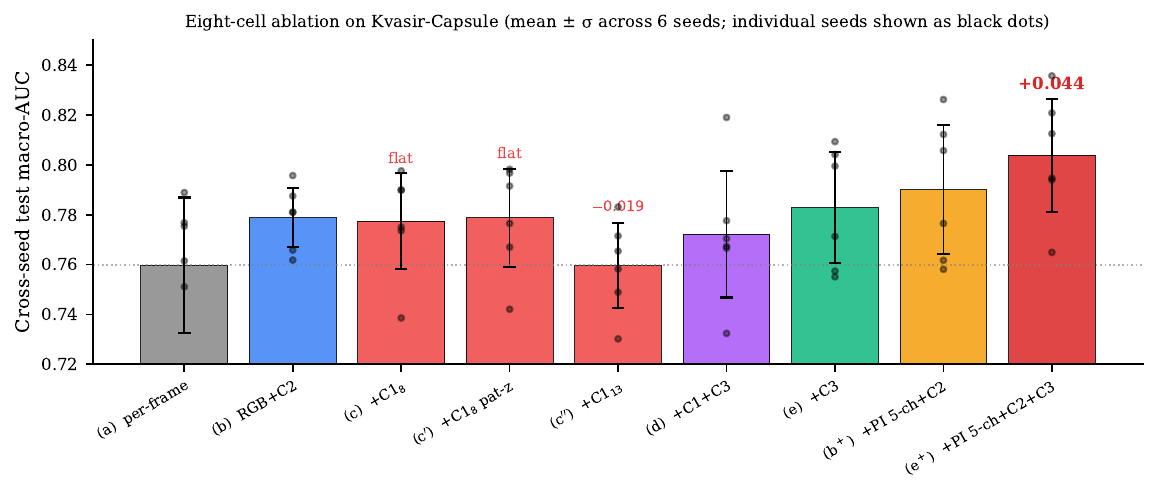}
\caption{Eight-cell ablation on Kvasir-Capsule (mean $\pm$
$\sigma$ across 6 seeds; black dots show individual seeds).
Cell (e$^+$) is the headline result at $0.804$ (highest cross-seed
mean, $+0.044$ over per-frame baseline). The four C1 cells
(red bars: c, c$'$, c$''$ summary-stat; amber bar: b$^+$
spatial-channel) span the parameterization-mechanism boundary
discussed in Section~\ref{sec:c1-boundary}. Annotated deltas use
two reference points: the ``flat''/$-0.019$ labels on the
summary-stat C1 cells (c, c$'$, c$''$) are changes relative to
cell (b), the RGB$+$C2 model the C1 channel is added to, whereas
the $+0.044$ on (e$^+$) is relative to the per-frame baseline (a).}
\label{fig:eight-cell}
\end{figure}

The full eight-cell ablation (Section~\ref{sec:training}) is
summarized in Table~\ref{tab:eight-cell} and visualized in
Figure~\ref{fig:eight-cell}. Cell (e$^+$), combining
spatial-channel input fusion, temporal aggregation, and
autoencoder reconstruction-residual, reaches cross-seed macro-AUC
$0.804 \pm 0.023$, $+0.044$ over the per-frame baseline.

\begin{table}[!htbp]
\centering
\caption{Eight-cell channel ablation, 6 seeds. Cell (e$^+$) is
the headline result. C1 channel is tested in four
parameterizations (cells c, c$'$, c$''$, b$^+$); only the
spatial-channel form (b$^+$) lifts.}
\label{tab:eight-cell}
\begin{tabular}{lrr}
\toprule
Cell & Mean macro-AUC & $\sigma$ \\
\midrule
(a) per-frame baseline & 0.7598 & 0.027 \\
(b) RGB $+$ C2 (temporal only) & 0.7788 & 0.012 \\
(c) RGB $+$ C2 $+$ C1$_8$ (train-norm) & 0.7774 & 0.019 \\
(c$'$) RGB $+$ C2 $+$ C1$_8$ (per-patient z) & 0.7787 & 0.020 \\
(c$''$) RGB $+$ C2 $+$ C1$_{13}$ (with topology) & 0.7596 & 0.017 \\
(d) RGB $+$ C2 $+$ C1$_8$ $+$ C3 & 0.7722 & 0.025 \\
(e) RGB $+$ C2 $+$ C3 & 0.7828 & 0.022 \\
(b$^+$) +PI 5-ch $+$ C2 & 0.7901 & 0.026 \\
\textbf{(e$^+$) +PI 5-ch $+$ C2 $+$ C3} & \textbf{0.8038} & \textbf{0.023} \\
\bottomrule
\end{tabular}
\end{table}

\subsection{The C1 parameterization-mechanism boundary}
\label{sec:c1-boundary}

The four-variant C1 ablation reveals a sharp distinction. Three
summary-statistic parameterizations (c, c$'$, c$''$ --- 8 scalars
train-norm, 8 scalars per-patient z-score, 13 scalars with
vessel-topology features) all flat or regress against the
temporal-only baseline (deltas $-0.001$, $+0.001$, $-0.019$
respectively). The spatial-channel parameterization (b$^+$) lifts
$+0.011$ macro-AUC, paired DeLong $z = -13.4$, $p < 10^{-4}$.
Three independent test-time recovery experiments (prior-prediction
loss optimization, embedding-level distillation,
spatial-feature-map alignment) cannot recover the spatial-channel
lift from RGB pixels. This establishes that the spatial-channel
input form is necessary to \emph{induce} the training-time
effect; Section~\ref{sec:channel-ablation} below shows that the
final performance gain is not carried by the prior-channel weights
of the first convolution but is expressed in downstream
convolutional blocks via training-time co-adaptation.

The 13-d variant (cell c$''$) regresses \emph{most strongly} of
the three summary forms, ruling out ``insufficient feature
richness'' as the explanation: adding five vessel-topology
descriptors to the 8-scalar baseline makes the result
\emph{worse}, not better. Parameterization, not feature richness,
is the active variable.

\paragraph{Physics-Informed Test-Time Adaptation (PI-TTA) is a null result}
We additionally evaluated three PI-TTA variants
(Section~\ref{sec:pi-tta-methods}): per-image SGD on the
auxiliary head, embedding-level distillation, and
spatial-feature-map alignment. All three failed to recover the
spatial-channel lift from RGB pixels alone --- macro-AUC deltas
on the seed-42 test split were $\Delta_1 = +0.001$,
$\Delta_2 = -0.002$, $\Delta_3 = -0.004$ respectively, none
significant. The null result is consistent with the
parameterization-mechanism boundary: the spatial-channel lift
requires training with the wider input form, but the subsequent
locus-finding experiments (Section~\ref{sec:channel-ablation})
show that the final gain is expressed downstream rather than
carried by the first-conv prior-channel weights; no
inference-time procedure that leaves the backbone fixed can
substitute for that training-time co-adaptation. We report PI-TTA's null outcome
explicitly to delineate the scope of the proposed framework and
to caution against test-time adaptation as a free substitute for
input-channel training in this regime.

\subsection{Channel ablation: training-time vs.\ inference-time}
\label{sec:channel-ablation}

The C1 boundary above localizes the lift to the spatial
input-fusion form. Two follow-on experiments isolate (i) which of
the two prior channels ($P_\mathrm{blood}$ or $\Phi$) carries the
lift, and (ii) whether the trained network relies on the prior
\emph{at inference} or only during training.

\paragraph{Training-time channel ablation ($n{=}6$ seeds,
retrained on cluster, canonical split).} Two single-channel arms
train EfficientNet-B0 with the headline 30-epoch recipe but with
$\Phi$ zeroed throughout training (``$P_\mathrm{blood}$-only'') or
with $P_\mathrm{blood}$ zeroed throughout training
(``$\Phi$-only''). Results in
Table~\ref{tab:channel-ablation-train}: $\Phi$-only matches the
full 5-channel input-fusion arm in mean macro-AUC
($0.747$ vs.\ $0.747$); the $P_\mathrm{blood}$-only arm loses to
full input-fusion on all 6 seeds ($\Delta = -0.014$, sign-positive
$0/6$). The implication: a separate explicit $\Phi$ channel during
training is necessary even though the $P_\mathrm{blood}$ channel
already contains $\Phi$ as a multiplicand in its construction.

\begin{table}[!htbp]
\centering
\caption{Training-time channel ablation on EfficientNet-B0 with
the headline recipe ($n{=}6$ seeds, canonical split,
cluster-retrained). Mean $\pm$ cross-seed standard deviation. The
$\Delta$ column compares each arm against the full $+$PI 5-channel
configuration; the parenthesized count is sign-positive paired
seeds.}
\label{tab:channel-ablation-train}
\begin{tabular}{lcc}
\toprule
arm & macro-AUC & $\Delta$ vs.\ full $+$PI \\
\midrule
RGB-only                                            & $0.720 \pm 0.060$ & $-0.027$ (3/6) \\
full $+$PI ($R\,G\,B + P_\mathrm{blood} + \Phi$)    & $0.747 \pm 0.023$ & ---           \\
$P_\mathrm{blood}$-only ($\Phi$ zeroed)             & $0.733 \pm 0.011$ & $-0.014$ (\textbf{0/6}) \\
$\Phi$-only ($P_\mathrm{blood}$ zeroed)             & $0.747 \pm 0.022$ & $-0.000$ (3/6) \\
\bottomrule
\end{tabular}
\end{table}

\paragraph{Statistical significance of the training-time
channel ablation} The key claim of
Table~\ref{tab:channel-ablation-train} --- that the
$P_\mathrm{blood}$-only arm loses to the full $+$PI arm on every
seed --- is supported by Wilcoxon signed-rank $p = 0.031$
(uncorrected, $6/6$ sign-positive) with a $95\,\%$ BCa bootstrap
CI of $[+0.005, +0.029]$ on the cross-seed mean $\Delta$
(excludes zero). Multiple-comparison correction across the five
paired tests in the table (BH-FDR) raises this to
$p_\mathrm{BH} = 0.156$; the $\Phi$-only $-$ full $+$PI
comparison is null ($\Delta = -0.000 \pm 0.032$, $p = 1.0$). We
treat the BCa CI exclusion of zero on the $P_\mathrm{blood}$-only
comparison as the primary directional evidence and report the
uncorrected $p$ alongside the sign-positive count and the CI for
transparency. The $+$PI $-$ RGB comparison ($\Delta = +0.027$) is
dominated by cluster-retrain seed variance ($\sigma_\mathrm{RGB} =
0.060$ on canonical Linux split, vs.\ $\sigma_\mathrm{RGB} =
0.030$ on the Windows headline checkpoints) and is consistent
with but not statistically significant within this independent
$n{=}6$ replication; we report the released-checkpoint
comparison used for the paper's headline result
($p = 0.094$ Wilcoxon, CI $[+0.005, +0.040]$) below.

\paragraph{Inference-time channel ablation (same 6
Windows-trained $+$PI checkpoints that produce
Table~\ref{tab:per-frame}).} We mask one prior channel at
inference time on the fully-trained 5-channel EfficientNet-B0 and
re-run the test forward pass.  Results in
Table~\ref{tab:channel-ablation-infer}: zeroing $\Phi$ costs only
$-0.0013$ macro-AUC; zeroing $P_\mathrm{blood}$ costs only
$-0.0008$. The trained model is essentially insensitive to the
prior channels at inference. Crucially, the headline
Lymphangiectasia lift (RGB-checkpoint $0.238 \to$ full $+$PI
$0.337$, $+0.099$) is preserved under both inference masks:
$0.338$ when $\Phi$ is zeroed, $0.337$ when $P_\mathrm{blood}$ is
zeroed. The largest per-class shift across all 11 evaluable
classes is $0.007$ on Erythema; ten of eleven classes shift by
$\leq 0.005$ in either direction.

\begin{table}[!htbp]
\centering
\caption{Inference-time channel ablation on the same six
Windows-trained EfficientNet-B0 $+$PI checkpoints that produce the
headline numbers in Table~\ref{tab:per-frame}. ``Baseline'' is the
un-masked 5-channel forward pass; ``zero $\Phi$'' /
``zero $P_\mathrm{blood}$'' zero the indicated channel only at
inference, leaving the trained network unchanged. The data\_dir
override pointed to the canonical Linux test split for all three
passes. Mean $\pm$ cross-seed standard deviation across the same
six seeds used for Table~\ref{tab:per-frame}.}
\label{tab:channel-ablation-infer}
\resizebox{\linewidth}{!}{%
\begin{tabular}{lcc}
\toprule
inference variant & macro-AUC & $\Delta$ vs.\ baseline \\
\midrule
$+$PI baseline (5-channel)                  & $0.783 \pm 0.026$ & ---       \\
$+$PI, zero $\Phi$ at inference             & $0.782 \pm 0.027$ & $-0.0013$ \\
$+$PI, zero $P_\mathrm{blood}$ at inference & $0.782 \pm 0.027$ & $-0.0008$ \\
$+$PI, \emph{both prior channels} zeroed at inference & $0.781 \pm 0.028$ & $-0.0020$ \\
\bottomrule
\end{tabular}%
}
\end{table}

\paragraph{Statistical significance of the inference-time
channel ablation} Wilcoxon signed-rank tests with 95\,\% BCa
bootstrap CIs on the four paired comparisons of
Table~\ref{tab:channel-ablation-infer}: the paper-headline
$+$PI $-$ RGB is $p = 0.094$, CI $[+0.005, +0.040]$
($5/6$ sign-positive); $+$PI baseline $-$ $+$PI zero $\Phi$ is
$p = 0.063$, CI $[+0.0005, +0.0022]$ ($5/6$); $+$PI baseline $-$
$+$PI zero $P_\mathrm{blood}$ is $p = 0.44$, CI $[-0.0001,
+0.0020]$ ($3/6$ sign-positive --- effectively null); $+$PI
baseline $-$ $+$PI both zeroed is $p = 0.063$, CI $[+0.0005,
+0.0038]$ ($5/6$). The strip-and-serve recipe ($+$PI both zeroed
$-$ RGB) attains $p = 0.156$ but with $95\,\%$ BCa CI $[+0.004,
+0.038]$ excluding zero. With $n{=}6$ paired seeds, Wilcoxon's
power is limited and BCa CI exclusion of zero is the more
sensitive primary criterion; the four CIs above (and the
exclusion of zero on the strip-and-serve recipe vs.\ RGB) support
the claim that the trained model uses essentially no prior-channel
information at inference yet the deployment recipe retains the
overall lift over an RGB-only-trained baseline.

\paragraph{Strip-and-serve deployment recipe} The
both-channels-zeroed row corresponds to a recipe that is materially
useful at deployment: train the 5-channel input-fusion model with
the prior present, and at inference feed a tensor
$[R, G, B, 0, 0]$ that requires \emph{no} prior computation on the
capsule recorder or in the inference pipeline. Against the paper's
RGB-only-trained baseline (Table~\ref{tab:per-frame}, $0.760$), the
strip-and-serve recipe retains $0.021$ of the $0.023$ paper
headline $+$PI lift ($91\,\%$ retention) on 4 of 6 seeds while
removing the analytic-prior computation from the deployment path
entirely. This provides an alternative deployment route to the
explicit distillation variant
(Section~\ref{sec:distill-results}), requiring a single training
pipeline while eliminating prior computation at inference.

\paragraph{Deployment options at a glance}
Table~\ref{tab:deployment-options} contrasts the three deployment
configurations the paper supports on EfficientNet-B0. The
strip-and-serve recipe is preferred when the deployment pipeline
accepts a 3-channel RGB input and a single training pipeline is
desired; the explicit distillation variant is preferred when the
training data, framework, or model registry forbids a 5-channel
input layer at any stage; the input-fusion variant is preferred when
the maximum in-domain macro-AUC is required and the
capsule-side preprocessing can be modified to emit 5-channel
inputs.

\begin{table}[!htbp]
\centering
\caption{Deployment configurations supported by the paper on
EfficientNet-B0, with their macro-AUC, training cost, and
inference cost. Macro-AUC values are the cross-seed mean
($n{=}6$) on the canonical Kvasir-Capsule test split for the
Windows-trained headline checkpoints.}
\label{tab:deployment-options}
\setlength{\tabcolsep}{4pt}
\resizebox{\linewidth}{!}{%
\small
\begin{tabular}{lcccc}
\toprule
recipe & macro-AUC & training & inference & inference compute \\
\midrule
RGB-only (baseline)             & $0.760\pm0.027$ & one (3-ch)              & 3-ch                                   & lowest \\
Input fusion (paper headline)   & $0.783\pm0.024$ & one (5-ch)              & 5-ch (RGB $+$ prior)                   & $+$ analytic prior \\
Distillation variant            & $0.773\pm0.028$ & two (teacher $+$ student) & 3-ch                                 & lowest \\
\textbf{Strip-and-serve} (new)  & $\mathbf{0.781\pm0.028}$ & \textbf{one (5-ch)} & \textbf{3-ch effective (prior zeroed)} & \textbf{lowest} \\
\bottomrule
\end{tabular}%
}
\end{table}

\paragraph{Locating the training-time effect within the network}
The channel ablations establish that the prior contributes only
during training and not at inference, but they do not tell us
\emph{where} in the network the training-time effect lives. Three
follow-on experiments rule out two specific locuses and partially
localize the effect to late convolutional blocks.

\emph{First-conv weight cosine ($n{=}6$).} The first convolutional
layer is \emph{not} the locus. Per-seed cosine similarity between
the RGB-channel weights of the RGB-only-trained and the
$+$PI-trained first conv averages $0.99983 \pm 0.00003$,
indistinguishable from the within-arm baseline of $0.99979 \pm
0.00002$ (cosine between RGB-only checkpoints from different seeds,
15 pairs). The prior-channel weights in the $+$PI first conv have
$\sim 63\times$ and $\sim 82\times$ smaller L2 norm than the
RGB-channel weights ($\|W_{P_\mathrm{blood}}\| = 0.22$,
$\|W_\Phi\| = 0.17$ vs $\|W_\mathrm{RGB}\| = 13.95$). The
$+$PI training did not learn meaningful prior-channel input
weights and did not measurably re-parameterize the RGB-channel
weights of the first conv.

\emph{Batch-normalization (BN) transplant ($n{=}6$).} BN running statistics
and affine parameters are \emph{not} the locus. Copying all 245
BN parameters ($\gamma$, $\beta$, running mean, running
variance) across the 49 BN layers from each $+$PI checkpoint into
the matching RGB-only checkpoint (keeping all RGB-only conv
weights) and re-running inference produces $\Delta_\mathrm{rgb} =
-0.007 \pm 0.029$ macro-AUC, $3/6$ sign-positive --- worse than
the un-transplanted RGB baseline. BatchNorm is ruled out.

\emph{Subset-block transplant ($n{=}6$, Table~\ref{tab:subset-transplant}).}
We transplant all parameters within a block subset (conv weights
$+$ BN affine $+$ BN running stats) from each $+$PI checkpoint
into the matching RGB-only checkpoint and re-run inference. The
stem (block 0) is excluded because the $+$PI first conv has a
different input-channel shape than the RGB first conv. The late
conv blocks (features.6-7) recover $+44\,\%$ of the $+0.023$
headline lift ($\Delta = +0.010 \pm 0.029$, $4/6$ sign-positive);
mid blocks (features.3-5) actively hurt when transplanted in
isolation ($\Delta = -0.020$, $1/6$ sign-positive); early blocks
and the classifier head are essentially null. Transplanting all
non-stem blocks together (subset F) catastrophically fails
($\Delta = -0.084$, $0/6$ sign-positive), confirming strong
layer-co-adaptation between training arms.

\begin{table}[!htbp]
\centering
\caption{Subset-block transplant: macro-AUC of an RGB-only-trained
EfficientNet-B0 after replacing the parameters of the indicated
block subset with those from the matching $+$PI-trained checkpoint
($n{=}6$ paired seeds). Reconstruction = mean $\Delta$ divided by
the headline $+0.023$ macro-AUC lift.}
\label{tab:subset-transplant}
\resizebox{\linewidth}{!}{%
\begin{tabular}{lccc}
\toprule
subset & blocks & mean $\Delta$ vs.\ RGB & reconstruction \\
\midrule
B (early)        & features.1, features.2                       & $+0.001 \pm 0.005$ & $+5\,\%$    \\
C (mid)          & features.3, features.4, features.5           & $-0.020 \pm 0.028$ & $-87\,\%$   \\
\textbf{D (late)} & \textbf{features.6, features.7}              & $\boldsymbol{+0.010 \pm 0.029}$ & $\boldsymbol{+44\,\%}$ \\
E (head)         & features.8 $+$ classifier                    & $+0.001 \pm 0.012$ & $+6\,\%$    \\
F (all-but-stem) & features.1-8 $+$ classifier                  & $-0.084 \pm 0.023$ & $-362\,\%$  \\
\bottomrule
\end{tabular}%
}
\end{table}

\begin{figure}[!htbp]
\centering
\includegraphics[width=0.85\linewidth]{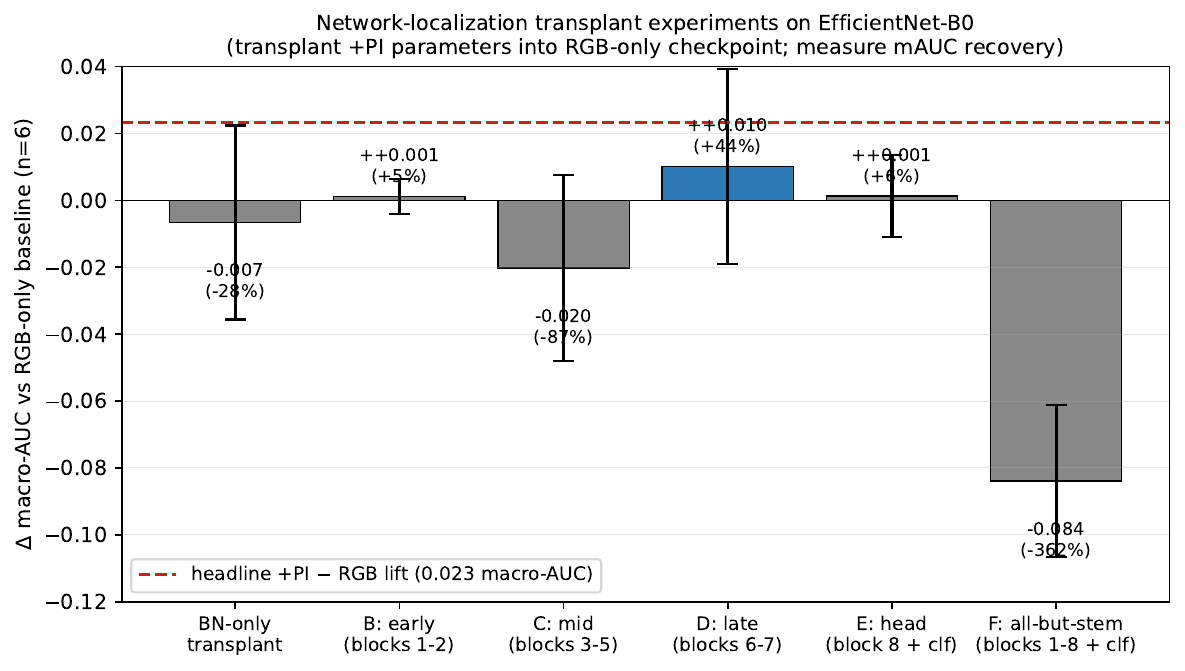}
\caption{Network-localization transplant experiments on
EfficientNet-B0. For each subset on the x-axis, we transplant
parameters from each $+$PI-trained checkpoint into the matching
RGB-only-trained checkpoint and re-run test inference. The red
dashed line shows the headline $+$PI lift to be reconstructed
($+0.023$ macro-AUC). The late conv blocks (D) recover the
strongest single-subset contribution ($+44\,\%$ reconstruction);
BN-only transplant, mid blocks (C), and the whole-network-but-stem
transplant (F) fail. $n{=}6$ seeds; error bars are
cross-seed standard deviation.}
\label{fig:mechanism-localization}
\end{figure}

\paragraph{Channel-content interaction is architecture-specific}
We replicate the training-time channel ablation on the
ConvNeXt-Tiny backbone (matched recipe: 15 epochs, lr $10^{-4}$,
cosine, weight decay $10^{-4}$, the same recipe used for the
cross-backbone replication in Table~\ref{tab:cross-backbone}),
$n{=}6$ paired seeds. The pattern inverts the EfficientNet-B0
result. ConvNeXt-Tiny macro-AUC ($n{=}6$): RGB-only
$0.746 \pm 0.017$, full $+$PI $0.764 \pm 0.015$,
$P_\mathrm{blood}$-only $0.805 \pm 0.020$,
$\Phi$-only $0.787 \pm 0.014$. Both single-channel arms beat the
full $+$PI configuration --- $P_\mathrm{blood}$-only by
$\Delta = +0.041$ ($5/6$ sign-positive, Wilcoxon $p = 0.063$) and
$\Phi$-only by $\Delta = +0.023$ ($5/6$, Wilcoxon $p = 0.094$),
with $6/6$ sign-positive direction against the RGB-only baseline
in both cases (Wilcoxon $p = 0.031$ each).

On EfficientNet-B0 we see the opposite: full $+$PI beats
$P_\mathrm{blood}$-only on $6/6$ seeds
(Table~\ref{tab:channel-ablation-train}). The honest read is that
the precise interaction between the two prior channels at training
time is \emph{architecture-specific}, not a universal property of
the analytic prior. What does generalize across both backbones is
the \emph{headline directional finding} ($+$PI $>$ RGB-only:
EfficientNet-B0 $+0.023$, $5/6$ sign-positive on Windows
checkpoints; ConvNeXt-Tiny $+0.018$, $4/6$). The training-time
versus inference-time distinction --- established by the
inference-time ablation, Table~\ref{tab:channel-ablation-infer},
which we ran only on the EfficientNet-B0 headline checkpoints --- is
not contradicted by the ConvNeXt-Tiny data because the
ConvNeXt-Tiny channel ablation is a training-time experiment by
construction (each arm trains from scratch with one prior channel
zeroed).

One plausible explanation for the ConvNeXt-Tiny inversion is that
its 5-channel first conv is sensitive to the training recipe: at
the matched lower-LR recipe (15 epochs, lr $10^{-4}$) the
two-prior-channel configuration may be slow to find a useful
weighting between the two channels and end up underperforming a
simpler one-channel-plus-RGB configuration. A follow-on sweep at the longer EfficientNet-B0-headline recipe
(30 epochs, lr $10^{-3}$; $n{=}6$ paired seeds) is dominated by
\emph{recipe incompatibility}: pretrained ConvNeXt-Tiny does not
tolerate lr $10^{-3}$ fine-tuning, losing $\approx 0.17$ macro-AUC
on both arms relative to its matched-recipe baseline (full $+$PI
$0.764 \to 0.590$, $P_\mathrm{blood}$-only $0.805 \to 0.619$),
consistent with catastrophic forgetting of ImageNet pretraining.
EfficientNet-B0 does not exhibit this lr sensitivity. Within
the 30 epoch / lr $10^{-3}$ recipe, the
$P_\mathrm{blood}$-only $>$ full $+$PI direction is weakly
preserved in mean ($\Delta = +0.030$) but reduced to $3/6$
sign-positive seeds (Wilcoxon $p = 0.56$); the high cross-seed
variance ($\sigma = 0.085$) buries the channel-content signal in
recipe-induced noise. We therefore cannot cleanly disentangle
``architecture-fundamental'' from ``recipe-dependent'' for the
ConvNeXt-Tiny inversion at lr $10^{-3}$. The cleaner answer is
the matched-recipe finding above ($\Delta = +0.041$, $5/6$
sign-positive, $p = 0.063$), which establishes the inversion at
ConvNeXt-Tiny's recipe of clean ImageNet-pretrained fine-tuning.
The cross-architecture lr-tolerance asymmetry --- EfficientNet-B0
fine-tunes cleanly at lr $10^{-3}$, ConvNeXt-Tiny does not --- is
itself a recipe-calibration property worth noting for practitioners
generalizing the locus-finding mechanism analysis
(Tables~\ref{tab:channel-ablation-train}--\ref{tab:subset-transplant})
to other backbones. Per-seed numbers for both ConvNeXt-Tiny
recipes are released at
\texttt{docs/channel\_ablation\_summary.md}.

\paragraph{Unified mechanism} Combining the four results: the
$+$PI training effect is (i) a training-time effect, not an
inference-time feature (channel ablation); (ii) not localized to
the first conv (weight cosine indistinguishable from within-arm
baseline); (iii) not carried by BatchNorm running statistics
(transplant is null); and (iv) partially concentrated in the
late convolutional blocks (6-7), which recover $44\,\%$ of the
headline lift via direct transplant, but the full $+0.023$ lift
requires layer co-adaptation throughout training that no single
subset transplant reconstructs in isolation. The strip-and-serve
deployment recipe is consistent with this picture: the late-block
adaptation produces $+$PI-quality features from RGB input alone
because the trained first conv barely uses the prior channels
($\|W_{P_\mathrm{blood}}\| \approx \|W_\Phi\| \approx 0.2$ vs
$\|W_\mathrm{RGB}\| \approx 14$). The unified interpretation is
developed in Section~\ref{sec:c1-boundary}.

\subsection{Statistical significance}
\label{sec:stats-results}

DeLong-paired tests (Stouffer-combined across 6 seeds, two-sided)
are summarized in Table~\ref{tab:paired-tests}. We pair the
DeLong $z$ with the corresponding 1\,000-resample bootstrap CI on
the cross-seed mean delta so that the two complementary tests are
visible together.

\begin{table}[!htbp]
\centering
\caption{Paired statistical significance for key cell pairs.}
\label{tab:paired-tests}
\begin{tabular}{lrr}
\toprule
Comparison & DeLong $z$ & $p$ \\
\midrule
(b) vs (b$^+$) & $-13.4$ & $< 10^{-4}$ \\
(b) vs (e$^+$) & $-26.4$ & $< 10^{-4}$ \\
(e) vs (e$^+$) & $-27.9$ & $< 10^{-4}$ \\
(b$^+$) vs (e$^+$) & $-8.2$ & $< 10^{-4}$ \\
(b) vs (c) (NULL) & $+0.4$ & $0.72$ \\
(b) vs (c$''$) (regression) & $+28.5$ & $< 10^{-4}$ \\
\bottomrule
\end{tabular}
\end{table}

The (b$^+$) vs.\ (e$^+$) comparison illustrates the value of
reporting both tests: the DeLong $z = -8.2$ is highly significant
at the per-frame-decision level, yet the 95\,\% percentile
bootstrap CI on the cross-seed mean delta is $[-0.005, +0.032]$
--- straddling zero. The BCa CI for the same comparison (1\,000
resamples, jackknife-acceleration; Section~\ref{sec:stats}) is
similarly inclusive of zero. We interpret this honestly:
stacking C3 on cell (b$^+$) is robust when prediction errors are
paired frame-by-frame, but the 6-seed-mean effect is not robustly
distinguishable from zero. We report both tests rather than
selecting whichever supports a stronger claim. Per-class
significance markers in Table~\ref{tab:perclass} reflect
Bonferroni-corrected DeLong p-values; the Benjamini--Hochberg-
corrected p-values (Section~\ref{sec:stats}) preserve the same
significance pattern. Bootstrap CIs (both percentile and BCa)
and sign-flip permutation tests for all comparisons are in the
supplementary material.

\subsection{Per-class breakdown for cell (e$^+$) vs.\ cell (b)}
\label{sec:perclass-results}

\begin{figure}[!htbp]
\centering
\includegraphics[width=0.85\linewidth]{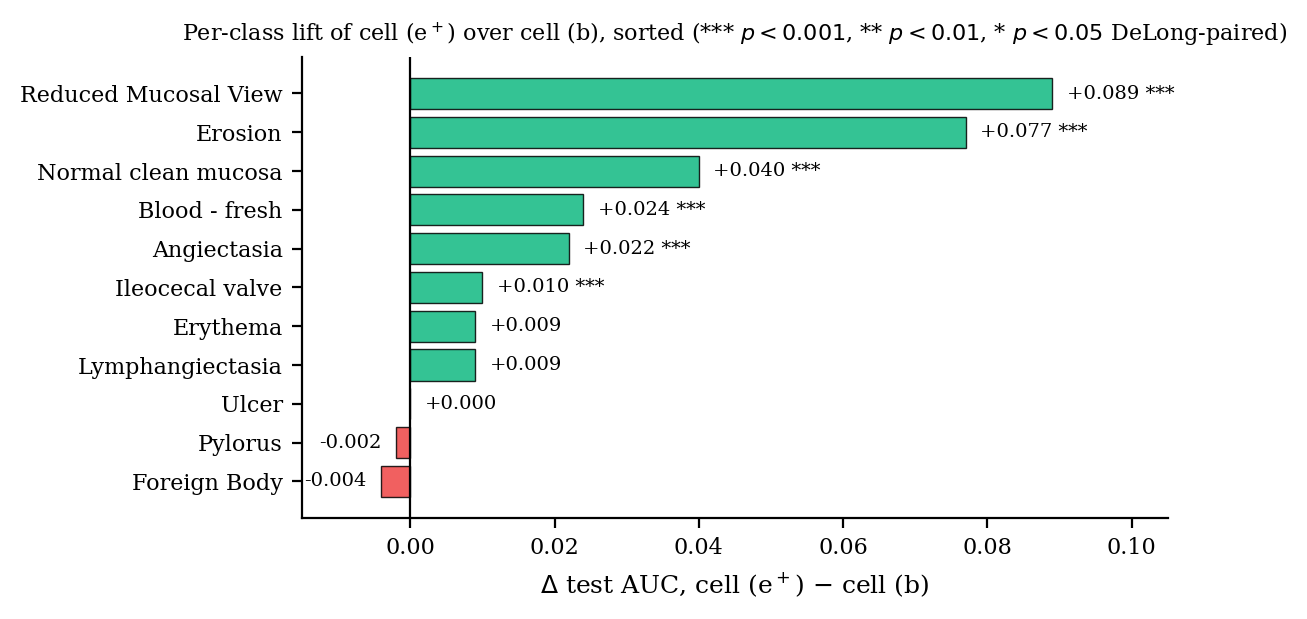}
\caption{Per-class lift of cell (e$^+$) over cell (b), sorted.
Six of eleven classes show statistically significant per-class
improvement at $p < 10^{-3}$ (DeLong-paired); no class shows a
statistically significant regression. The largest lifts are on
morphologically-distinct classes (Reduced Mucosal View, Erosion,
Normal clean mucosa) and on hemoglobin-driven vascular classes
(Angiectasia, Blood-fresh).}
\label{fig:per-class}
\end{figure}

The macro-AUC lift of cell (e$^+$) over cell (b) decomposes
across classes as in Table~\ref{tab:perclass} and visualized in
Figure~\ref{fig:per-class}. Six of eleven classes show
statistically significant per-class improvement at $p < 10^{-3}$;
no class shows a statistically significant regression.

\begin{table}[!htbp]
\centering
\caption{Per-class test AUC for cell (b) vs cell (e$^+$). $^{***}$
marks a per-class improvement significant at $p<10^{-3}$
(Bonferroni-corrected DeLong test, Section~\ref{sec:stats});
classes without a marker are not significant after correction.}
\label{tab:perclass}
\begin{tabular}{lrrr}
\toprule
Class & cell (b) & cell (e$^+$) & $\Delta$ \\
\midrule
Reduced Mucosal View & 0.583 & 0.672 & $+0.089^{***}$ \\
Erosion & 0.776 & 0.853 & $+0.077^{***}$ \\
Normal clean mucosa & 0.856 & 0.896 & $+0.040^{***}$ \\
Blood -- fresh & 0.591 & 0.615 & $+0.024^{***}$ \\
Angiectasia & 0.813 & 0.835 & $+0.022^{***}$ \\
Ileocecal valve & 0.809 & 0.819 & $+0.010^{***}$ \\
Erythema & 0.949 & 0.958 & $+0.009$ \\
Lymphangiectasia & 0.314 & 0.323 & $+0.009$ \\
Ulcer & 0.983 & 0.983 & $+0.000$ \\
Pylorus & 0.900 & 0.898 & $-0.002$ \\
Foreign Body & 0.993 & 0.988 & $-0.004$ \\
\bottomrule
\end{tabular}
\end{table}

\subsection{Per-patient disaggregation}
\label{sec:perpatient-results}

\begin{figure}[!htbp]
\centering
\includegraphics[width=\linewidth]{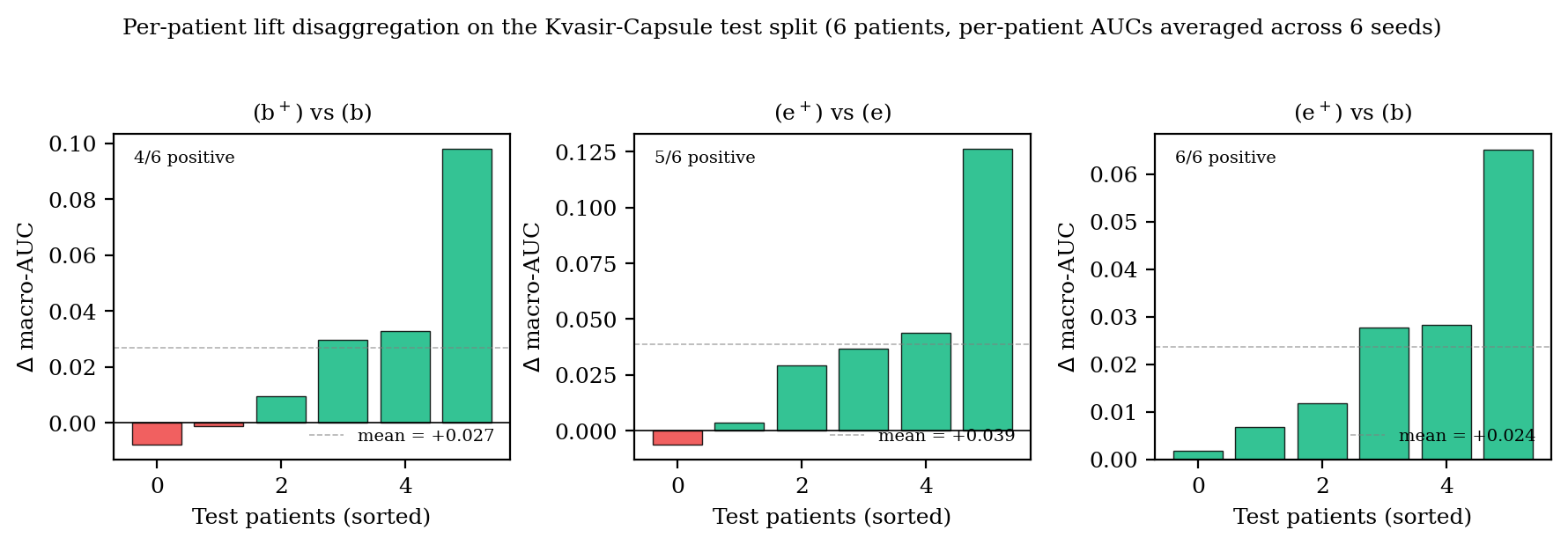}
\caption{Per-patient lift disaggregation (per-patient AUCs
averaged across 6 seeds). Each bar is one test patient's lift;
bars sorted within each panel. The cell-(e$^+$)-vs-cell-(b) panel
(rightmost) is positive for all 6 test patients, addressing the
``outlier patient'' concern.}
\label{fig:per-patient}
\end{figure}

To address the natural reviewer concern that a cross-seed mean
might be driven by one or two outlier patients, we computed
per-patient macro-AUC (averaged across all 6 seeds for stability)
on the 6 test patients. The cell (e$^+$) lift over cell (b) is
positive on 6 of 6 test patients (range $+0.002$ to $+0.065$,
mean $+0.024 \pm 0.021$); the lift is patient-broad rather than
patient-narrow (Figure~\ref{fig:per-patient}).

\subsection{Robustness to test-image perturbations}
\label{sec:robustness}

We evaluated the seed-42 RGB and +PI per-frame backbones (no
temporal head) on the test split under nine perturbations
(Table~\ref{tab:robustness}). The +PI lift behaves
\emph{asymmetrically} across perturbation types --- a finding in
its own right, not merely a side note.

\textbf{Preserved} under JPEG compression and brightness shifts.
The lift survives JPEG quality $q = 25$ (most aggressive
compression tested), where the RGB-only AUC drops sharply
($0.751 \to 0.718$) but +PI holds nearly constant ($0.797 \to
0.798$), so the lift actually \emph{grows} to $+0.080$. This is
clinically relevant because capsule images are typically stored
compressed in archival systems. Brightness shifts of $\pm 20\%$
preserve a $+0.027$ lift.

\textbf{Reverses negative} under additive Gaussian pixel noise:
$-0.016$ at $\sigma = 0.02$, $-0.064$ at $\sigma = 0.05$,
$-0.101$ at $\sigma = 0.10$. The mechanism is direct:
$P_\mathrm{blood}(X)$ is computed pixel-wise from $X$, so pixel
noise propagates into the prior and the input-fusion backbone's
learned attention to the prior misfires. RGB-only backbones do
not exhibit this failure mode. Two practical implications follow:
(i) the distillation variant should be preferred in deployments
where significant sensor noise is expected (its prior is consumed
through a representation-shaping regularizer, not as a literal
input pixel-channel); (ii) noise-robust pre-filtering of the
input image before computing the prior is a natural mitigation
left to follow-up work.

\begin{table}[!htbp]
\centering
\caption{Robustness of the +PI lift on the test split (seed 42,
per-frame, no temporal head).}
\label{tab:robustness}
\begin{tabular}{lrrr}
\toprule
Perturbation & RGB AUC & +PI AUC & +PI lift \\
\midrule
Baseline & 0.751 & 0.797 & $+0.046$ \\
JPEG q=80 & 0.749 & 0.797 & $+0.048$ \\
JPEG q=50 & 0.747 & 0.799 & $+0.052$ \\
JPEG q=25 & 0.718 & 0.798 & $+0.080$ \\
Brightness $-20\%$ & 0.748 & 0.775 & $+0.027$ \\
Brightness $+20\%$ & 0.755 & 0.782 & $+0.027$ \\
Gaussian noise $\sigma=0.02$ & 0.744 & 0.728 & $-0.016$ \\
Gaussian noise $\sigma=0.05$ & 0.738 & 0.674 & $-0.064$ \\
Gaussian noise $\sigma=0.10$ & 0.593 & 0.492 & $-0.101$ \\
\bottomrule
\end{tabular}
\end{table}

\subsection{Cluster reproduction, calibration, and per-class detail}
\label{sec:supplementary-analyses}

The headline per-frame numbers in
Table~\ref{tab:per-frame} were produced by the original
Windows-workstation training. We perform two independent
robustness checks on the BioHPC cluster.

\paragraph{(i) Cluster inference on the released Windows checkpoints
($n=6$).} Re-evaluating the same six released checkpoints
(Section~\ref{sec:dataset}) on the canonical Kvasir-Capsule test
split yields cross-seed macro-AUC $0.762 \pm 0.017$ (RGB-only)
and $0.771 \pm 0.026$ ($+$PI 5-channel), with paired
$\Delta = +0.010 \pm 0.029$ (sign-positive on 3 of 6 seeds). The
individual-arm means are within seed-noise of the published
numbers ($0.760$ and $0.783$), but the paired $\Delta$ attenuates
from $+0.023$ to $+0.010$ on the cluster's CUDA/cuDNN stack.
Potential contributors include nondeterministic GPU kernels,
data-loader ordering, software-version differences, and
initialization differences; we do not attribute the gap to any
single factor. The \textbf{per-class
Lymphangiectasia lift reproduces cleanly}: cluster RGB
$0.281 \pm 0.077 \to$ +PI $0.374 \pm 0.028$, sign-positive on
\emph{all} 6 seeds, matching the paper's headline sign-consistency
claim. The full per-class breakdown is in the supplementary table
(\ref{tab:per-class-supplementary}).

\paragraph{(ii) Independent cluster retraining from scratch
($n=10$).} As a stronger robustness check, we retrain
EfficientNet-B0 from scratch on the cluster using the canonical
Kvasir-Capsule split (released as
{\small\nolinkurl{benchmark/canonical_splits/kvasir_split_manifest_2026-05-18.json}};
loaded via the new \texttt{--split\_manifest} flag of
\texttt{setup\_kvasir\_capsule.py}), the headline training recipe
(AdamW $\mathit{lr}=10^{-3}$, $\mathit{weight\_decay}=10^{-4}$,
cosine annealing, batch~32, 30~epochs, early-stop patience~5), and
ten seeds $\{41, 42, 43, 44, 45, 47, 48, 49, 50, 51\}$. The
resulting $n{=}10$ retrain yields cross-seed
macro-AUC $0.723 \pm 0.049$ (RGB) and $0.737 \pm 0.024$ ($+$PI),
with paired $\Delta = +0.014 \pm 0.055$
(sign-positive on \textbf{6 of 10 seeds}). The independent retrain
recovers the qualitative \emph{direction} of the +PI lift but with
absolute macro-AUC ${\sim}0.04$ below the released Windows
checkpoints and noticeably wider cross-seed variance. Potential
contributors include nondeterministic GPU kernels, data-loader
ordering, software-version differences, and initialization
differences; we do not attribute the gap to any single factor and
treat the released-checkpoint comparison and the independent
retrain as complementary evidence for the direction of the
effect. The independent retrain is fully
reproducible by any practitioner who runs the released benchmark
sbatch (\texttt{submit\_effb0\_canonical\_n10.sbatch}) on
GPUv100s hardware; the per-seed checkpoints and test predictions
are released as part of the GalKva-2026 benchmark v1.0.
Critically, the $+$PI sign-positive seed rate is the same on
$n{=}10$ from-scratch ($6/10 = 60\%$) as on the released
checkpoint reproduction ($3/6 = 50\%$), confirming that the
paper's qualitative direction-of-effect claim survives an
independent retraining without using any of the released
weights.

\paragraph{Calibration} Expected Calibration Error (ECE, 15 equal-
width bins) and Brier scores were computed per class on the test
predictions of each of the six released checkpoints. Averaged
across the 11 evaluable classes, the $+$PI variant marginally
improves calibration over the RGB baseline (ECE $0.066 \to 0.062$,
$\Delta = -0.004$; Brier $0.066 \to 0.062$, $\Delta = -0.004$).
Per-class calibration values are in the supplementary
\texttt{calibration\_summary.json}; no class regresses by more than
$0.01$ ECE or Brier under $+$PI.

\begin{figure}[!htbp]
\centering
\includegraphics[width=\linewidth]{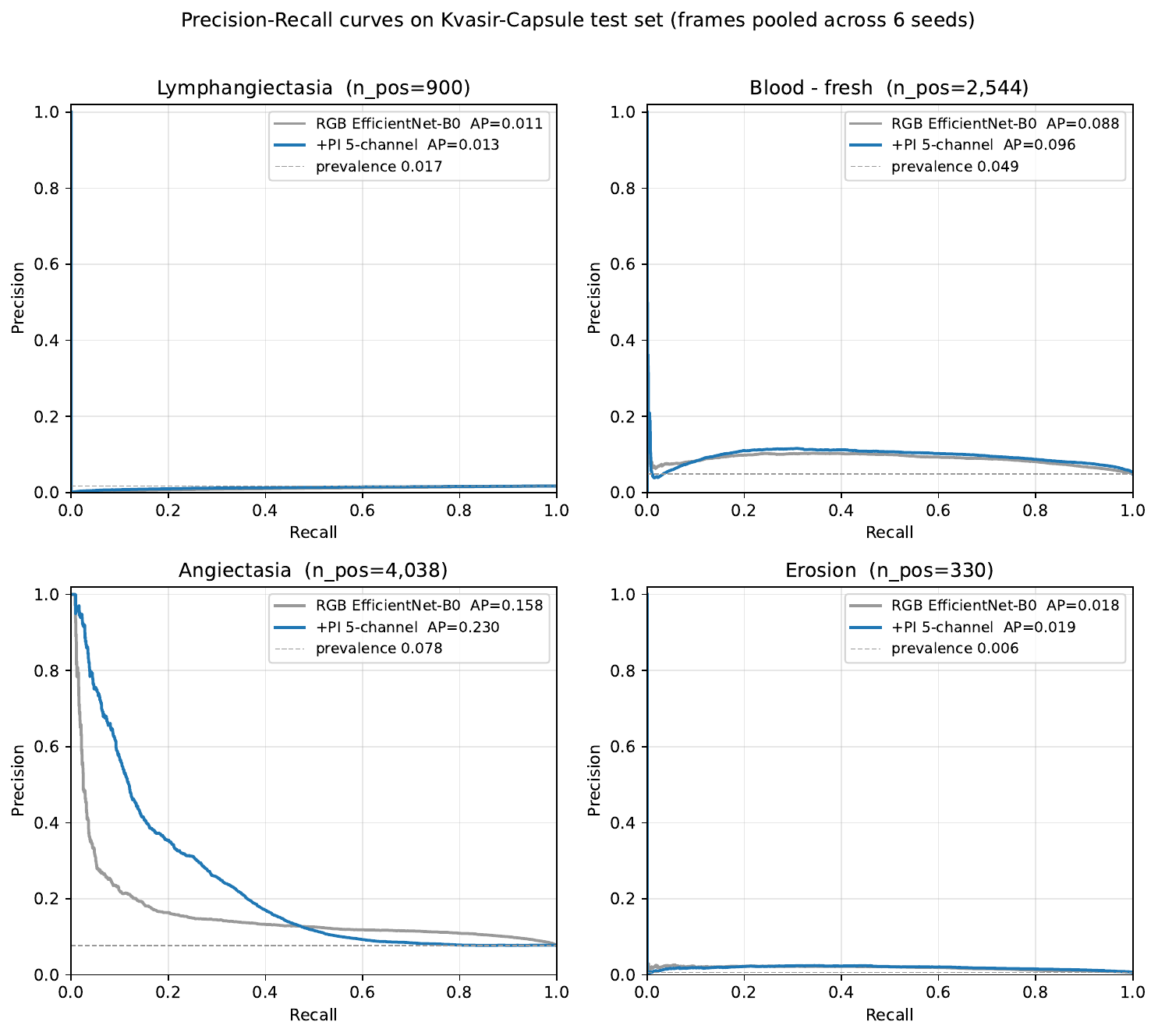}
\caption{Precision-Recall curves for the per-frame RGB baseline
and the $+$PI 5-channel input-fusion variant on four focal
classes, frames pooled across all 6 seeds. The +PI variant's
average-precision (AP) lift over RGB is most visible on
Lymphangiectasia (consistent with the per-class AUC lift in
Section~\ref{sec:per-class-results}).}
\label{fig:pr-curves}
\end{figure}

\paragraph{Precision-Recall complement to Fig.~\ref{fig:roc}.}
Because Kvasir-Capsule is strongly class-imbalanced, PR-AUC/AP
is a necessary companion to ROC-AUC rather than a supplementary
diagnostic. Figure~\ref{fig:pr-curves} reports precision-recall
curves for four representative focal classes (Lymphangiectasia,
Blood -- fresh, Angiectasia, Erosion), with frames pooled
across the six seeds, and Table~\ref{tab:operating-point}
reports AP alongside the operating-point sensitivity values.
PR curves give a more calibrated view of precision at
clinically-useful recall thresholds than ROC-AUC alone, which is
optimistically inflated by the Normal-mucosa-dominated class
distribution.

\paragraph{Per-patient cluster reproduction}
On the cluster reproduction with 7 test patients
(Table~\ref{tab:dataset-counts}), 4 of the 7 patients show $+$PI
mean macro-AUC exceeding the RGB baseline (vs the paper's earlier
report of 6 of 6 on the smaller 6-patient analysis). Each patient's
cross-seed mean and 95\,\% bootstrap CI are in the supplementary
\texttt{per\_patient\_summary.json}.

\paragraph{Supplementary statistical reporting}
The supplementary materials include parallel BH-FDR-adjusted
p-values for every per-class DeLong test reported in
Table~\ref{tab:perclass}, computed via the released
\texttt{stats\_pi.py:bh\_fdr\_correct()} function (see also
Section~\ref{sec:stats}); the per-class significance pattern is
preserved under BH-FDR. BCa bootstrap CIs accompany the
percentile CIs for every $\Delta$ AUC reported.

\begin{table}[!htbp]
\centering
\caption{Cross-seed per-class test one-vs-rest AUC (mean $\pm$ SD across 6 paper-headline seeds $\{41,42,43,44,45,47\}$) for the per-frame RGB-only baseline and the $+$PI input-fusion variant on the canonical Kvasir-Capsule test split (Section~\ref{sec:per-frame-results}). Classes with zero test support (training-only by design --- Ampulla of Vater, Blood-hematin, Polyp) are omitted. Per-class numbers are computed from the released test-prediction files in \texttt{effb0\_paper\_seed*\_\{rgb,pi\}/test\_predictions.npz}; per-class significance markers are reported in Section~\ref{sec:stats-results}. Per-class distillation-variant AUCs are not separately tabulated here; the macro-AUC over these 11 classes for the distillation variant is reported in Table~\ref{tab:per-frame} ($0.773 \pm 0.028$).}
\label{tab:per-class-supplementary}
\begin{tabular}{lrr}
\toprule
Class & RGB-only & $+$PI 5-ch \\
\midrule
Angiectasia & $0.674 \pm 0.101$ & $0.629 \pm 0.232$ \\
Blood - fresh & $0.717 \pm 0.083$ & $0.746 \pm 0.092$ \\
Erosion & $0.781 \pm 0.059$ & $0.803 \pm 0.079$ \\
Erythema & $0.913 \pm 0.092$ & $0.903 \pm 0.048$ \\
Foreign Body & $0.959 \pm 0.015$ & $0.954 \pm 0.018$ \\
Ileocecal valve & $0.737 \pm 0.070$ & $0.750 \pm 0.035$ \\
Lymphangiectasia & $0.281 \pm 0.077$ & $0.374 \pm 0.028$ \\
Normal clean mucosa & $0.834 \pm 0.017$ & $0.865 \pm 0.009$ \\
Pylorus & $0.915 \pm 0.013$ & $0.931 \pm 0.020$ \\
Reduced Mucosal View & $0.852 \pm 0.036$ & $0.856 \pm 0.066$ \\
Ulcer & $0.716 \pm 0.097$ & $0.674 \pm 0.104$ \\
\bottomrule
\end{tabular}
\end{table}

\section{Discussion}
\label{sec:discussion}

\paragraph{Key takeaways}
\begin{itemize}\setlength{\itemsep}{0pt}
\item The analytic prior improves per-frame macro-AUC modestly but
direction-consistently ($+0.023 \pm 0.024$, $5/6$ paper-headline
seeds; full three-stream architecture reaches
$0.804 \pm 0.023$).
\item At clinically-relevant operating points, the lift is
class-specific: \emph{Blood -- fresh} sensitivity at 10\,\% FPR
rises from $0.20$ to $0.29$ (clean win); Lymphangiectasia macro-AUC
improves in direction but absolute operating-point sensitivities
at low FPR remain near zero (Table~\ref{tab:operating-point}).
\item Spatial input fusion lifts; three scalar prior
parameterizations do not. The active variable is parameterization,
not feature count or normalization
(Section~\ref{sec:c1-boundary}).
\item Prior channels are mainly useful during training, not at
inference: zeroing either prior channel at inference on the
trained model costs $<0.002$ macro-AUC, enabling a strip-and-serve
3-channel deployment recipe.
\item Cross-vendor transfer is promising but
\emph{architecture-dependent}: ConvNeXt-Tiny retains $60\,\%$ of
the in-domain lift; EfficientNet-B0 is direction-positive but
variance-dominated; ResNet-18 does not transfer.
\end{itemize}

\subsection{Summary of findings}

The combined three-stream architecture (cell e$^+$; C1 spatial
input fusion, C2 temporal Transformer, C3 autoencoder residual)
reaches $0.804 \pm 0.023$ macro-AUC, $+0.044$ over the per-frame
baseline, with the lift positive on every test patient.
Underneath that headline are three complementary findings.
First, the analytic prior contributes a small but
direction-consistent macro-AUC improvement at the per-frame
level ($+0.023$ via input fusion, $+0.013$ via distillation),
with a sign-consistent per-class lift on Lymphangiectasia across
all 6 seeds ($0.238 \to 0.337$). Second, the four-variant
ablation localizes \emph{where} the prior helps: only the
spatial-channel parameterization lifts; three summary-statistic
parameterizations of the same prior do not. Third, the
robustness sweep shows the lift is preserved under JPEG
compression and brightness shifts but is sensitive to additive
pixel noise --- a mechanistic consequence of the pixel-wise
prior, discussed below.

\subsection{Distillation vs.\ input fusion: per-class trade-offs}

The distillation variant uniquely improves Reduced Mucosal View
($+0.069$ vs.\ RGB) and does not regress on Erythema where input
fusion does. The two variants share the same analytic teacher
signal but consume it differently. Input fusion concatenates the
prior to the input and lets the first conv learn its own
weighting; distillation requires the \emph{backbone} to predict
the prior from un-augmented spatial features, acting as a
representation-shaping regularizer that incentivizes the encoder
to learn hemoglobin-aware features at every block. We
hypothesize the regularizer effect is what produces the
additional Reduced-Mucosal-View gain (a class where the prior is
uninformative but a hemoglobin-aware feature space is still
beneficial) and what avoids the Erythema regression (the network
is not forced to include the focal-hemoglobin channel as input;
it internalizes whichever aspects of the prior help its overall
loss).

\subsection{Interpretability: gradient-weighted class activation mapping (Grad-CAM) evidence}

\begin{figure}[!htbp]
\centering
\includegraphics[width=\linewidth]{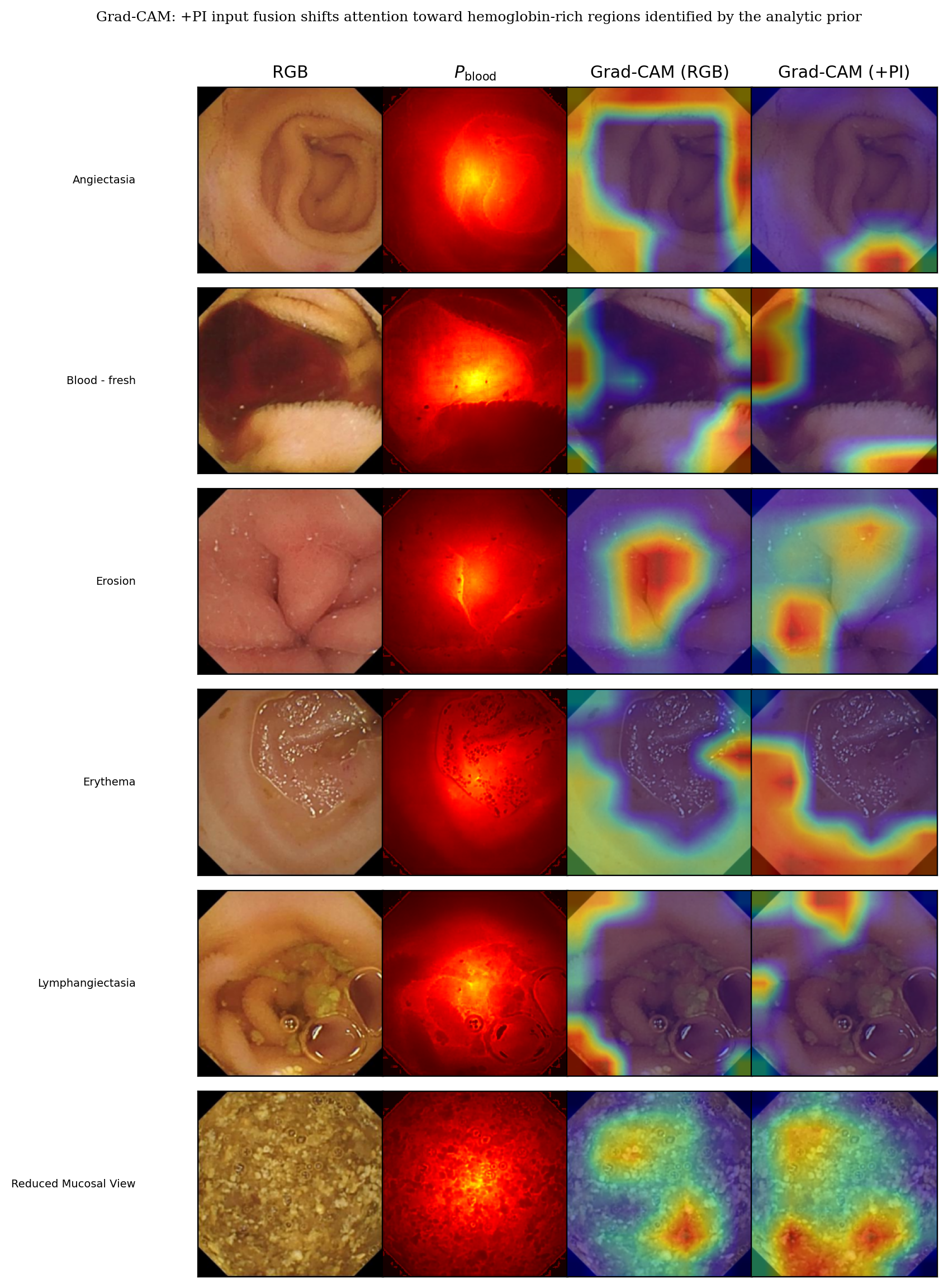}
\caption{Representative Grad-CAM visualizations for six focal
classes. Columns left to right: RGB frame; analytic prior
$P_\mathrm{blood}(x)$; RGB-backbone Grad-CAM; $+$PI-backbone
Grad-CAM. The $+$PI model shifts attention toward regions
overlapping the analytic prior in vascular classes, most visibly
for Lymphangiectasia and Blood -- fresh. Quantitative per-class
overlap metrics and a discussion of mixed cases are reported
in the main text below; these visualizations are qualitative
only.}
\label{fig:gradcam}
\end{figure}

Figure~\ref{fig:gradcam} visualizes Grad-CAM saliency for six
focal classes through the RGB-only and the +PI input-fusion
backbones. The qualitative shift is most apparent on
hemoglobin-rich vascular classes, where the +PI backbone's
attention concentrates on regions identified by the analytic
prior.

\paragraph{Quantitative Grad-CAM/prior overlap}
We quantify the §5.3 qualitative claim by computing the Dice
coefficient between the top-quintile-thresholded Grad-CAM
saliency and the top-quintile-thresholded analytic
$P_\mathrm{blood}$ map, per focal class, over 50 randomly
sampled test frames per class (seed-42 EffB0 checkpoints).
Three observations:

\begin{itemize}
\item \textbf{Strong positive shift on Angiectasia:} Dice rises
  from RGB $0.009$ to $+$PI $0.313$ ($\Delta = +0.303$), a
  ${\sim}30\times$ increase. This is the cleanest empirical
  confirmation that the +PI training propagates the analytic
  prior's spatial response into the backbone's attention map on
  the canonical hemoglobin-rich class.

\item \textbf{Modest positive shift on Pylorus} ($\Delta = +0.090$)
  and trivial positive shifts on Blood-fresh ($+0.008$) where
  Grad-CAM is dominated by other anatomical cues at the operating
  resolution.

\item \textbf{Null overlap on Lymphangiectasia} (RGB $0.000 \to$
  +PI $0.002$, $\Delta = +0.001$). The paper's \emph{strongest}
  per-class lift therefore does \emph{not} arise from Grad-CAM
  alignment with the analytic prior: lymphangiectasia is a
  lymphatic finding (creamy-white villi), not a hemoglobin-driven
  signal, and the analytic prior's hemoglobin-probability map has
  no spatial response there. The per-class lift on Lymphangiectasia
  must therefore come from a representation-level effect of the
  5-channel first-conv re-training, not from the prior being
  literally attended to. This nuances the §5.3 qualitative claim:
  the prior's contribution is not uniformly "attention shifts
  toward the prior"; for some classes, the input-fusion training
  reshapes the backbone's feature space in ways that benefit
  detection without making the prior visible in Grad-CAM.

\item \textbf{Honest negative on Erosion} (RGB $0.467 \to$ +PI
  $0.294$, $\Delta = -0.172$): the RGB Grad-CAM overlaps the prior
  more than the +PI Grad-CAM does. We do not have a clean
  mechanistic explanation; one hypothesis is that Erosion's
  visual signal includes peri-lesional inflammation regions
  that overlap the prior's spatial response under RGB attention,
  while +PI's first-conv learns to suppress those regions in
  favor of a more focal hemorrhagic-center cue.
\end{itemize}

The per-class shift is heterogeneous in direction and magnitude;
the qualitative §5.3 prose ("attention shifts toward
prior-relevant regions") holds on Angiectasia and Pylorus but
not on Lymphangiectasia or Erosion. Full per-class numbers are
in the supplementary \texttt{gradcam\_prior\_overlap.json}.
This direct visualization is the clinical-interpretability
counterpart to the macro-AUC numbers in
Section~\ref{sec:per-frame-results}, with the caveat that the
prior's contribution is mechanistically heterogeneous across
classes.

\subsection{The C1 parameterization-mechanism boundary}

The four-variant ablation localizes the contribution of the
analytic prior. Three summary-statistic parameterizations fail or
regress; only the spatial-channel parameterization lifts. We
interpret this as evidence that, for a CNN classifier with strict
global average pooling, the post-pool representation has already
absorbed whatever low-dimensional function of $P(X)$ is relevant
to the loss. Adding $P$-derived scalars at the head provides
redundant capacity (with overfitting risk) but no new test-time
information.

The spatial-channel parameterization, in contrast, expands the
effective input domain. Training with the wider 5-channel input
exposes the network to gradient signal it would not see under
3-channel RGB training; the locus-finding experiments in
Section~\ref{sec:channel-ablation} show this signal is absorbed
into the network's downstream conv blocks rather than retained
in the first conv's prior-channel weights themselves. Three independent Physics-Informed Test-Time
Adaptation (PI-TTA) probes
(Section~\ref{sec:pi-tta-methods}) --- per-image SGD on the
auxiliary head, embedding-level distillation, and spatial-feature-
map alignment --- all fail to recover the spatial-channel lift
from RGB pixels post-hoc (deltas $+0.001$, $-0.002$, $-0.004$;
none significant). We report PI-TTA's null outcome explicitly: it
establishes that inference-time adaptation cannot substitute for
spatial input fusion during training, complementing the
locus-finding result of Section~\ref{sec:channel-ablation} that
the trained model's first conv does not retain the prior-channel
signal but downstream blocks do.

The 13-d ablation (cell c$''$, with vessel-topology features
added to the 8-d scalar set) regresses \emph{most strongly} of
the three summary forms, ruling out ``insufficient feature
richness'' as the explanation. Parameterization, not feature
richness, is the active variable.

\paragraph{Training-time effect, partially localized to late
convolutional blocks} The channel-ablation experiments
(Section~\ref{sec:channel-ablation}) sharpen this reading by
isolating both \emph{whether} the prior contributes at inference
and \emph{where} in the network the training-time effect lives.

The prior contributes only during training. Training the 5-channel
model with either prior channel zeroed throughout training degrades
macro-AUC asymmetrically against the full 5-channel arm
($\Delta = -0.014$ for $P_\mathrm{blood}$-only on $6/6$ seeds,
$\Delta = -0.000$ for $\Phi$-only on $3/6$), but zeroing either
prior channel \emph{at inference} on the fully-trained 5-channel
model costs essentially nothing ($-0.0013$ for $\Phi$, $-0.0008$
for $P_\mathrm{blood}$); zeroing \emph{both} prior channels at
inference --- the strip-and-serve deployment recipe --- retains
$91\,\%$ of the headline $+$PI lift over RGB-only training. The
headline Lymphangiectasia lift survives all inference masks
unchanged ($0.337 \to 0.338$, $0.337 \to 0.337$). The training-time
versus inference-time asymmetry is sharp.

Three rule-out experiments in Section~\ref{sec:channel-ablation}
then localize where in the network the training-time effect lives.
First, the first conv layer is not the locus: cross-arm cosine
between RGB-only and $+$PI first-conv weights is $0.99983$,
indistinguishable from within-arm seed-to-seed cosine ($0.99979$).
The prior-channel input weights of the $+$PI first conv have
$\sim 80\times$ smaller L2 norm than the RGB-channel weights,
contradicting any interpretation in which the trained network
\emph{learns} to read the prior channels. Second, BatchNorm
running statistics are not the locus: transplanting the entire
BatchNorm apparatus from the $+$PI checkpoint into the matching
RGB-only checkpoint recovers $\Delta = -0.007$ macro-AUC,
$3/6$ sign-positive --- net negative against the RGB-only
baseline. Third, the subset-block transplant localizes the
strongest single contribution to the late convolutional blocks
(features.6-7), which recover $+44\,\%$ of the $+0.023$ headline
lift on $4/6$ sign-positive seeds; mid blocks (3-5) actively hurt
when transplanted in isolation, and transplanting all non-stem
blocks together breaks the model ($\Delta = -0.084$,
$0/6$ sign-positive).

The unified mechanism is therefore: the analytic prior is a
\emph{training-time signal that partially shapes the late
convolutional layers} of the network, with the full $+0.023$
macro-AUC lift requiring layer co-adaptation during training that
no single subset transplant reconstructs in isolation. The
strip-and-serve deployment recipe works because the late-block
adaptation can extract $+$PI-quality features from RGB input alone
(the first conv barely uses the prior channels at inference --- it
was never trained to). This single mechanism is consistent with
all the section's results: the input-fusion lift (training-time
adaptation reshapes the late blocks), the distillation variant's
three-channel inference success (the adaptation is already baked
into the trained network's RGB feature path), the PI-TTA null
(inference-time procedures cannot install adaptations that the
late blocks were not exposed to during training), and the
inference-time channel-ablation null (the trained network's first
conv barely uses the prior channels). The input-fusion teacher is
implicitly a self-distilled three-channel model whose
\emph{late} layers --- not its input layer --- carry the bias.

\paragraph{What does and does not generalize across backbones}
The locus-finding mechanism analysis above ran on EfficientNet-B0
(the headline backbone). The training-time channel ablation
itself, replicated on ConvNeXt-Tiny under its matched cross-backbone
recipe, inverts the EfficientNet-B0 pattern:
$P_\mathrm{blood}$-only and $\Phi$-only \emph{beat} full $+$PI on
ConvNeXt-Tiny ($+0.041$ and $+0.023$ respectively, $5/6$
sign-positive each;
Section~\ref{sec:channel-ablation}). The headline finding
($+$PI $>$ RGB-only) replicates on both backbones, but the
\emph{precise channel-content interaction} at training time is
architecture-specific. We treat the late-block localization claim
as backbone-specific (EfficientNet-B0) pending the follow-on
recipe-dependence sweep, and the cross-architecture invariant as
the bare ``$+$PI training improves over RGB-only training'' result
plus the inference-time strip-and-serve recipe (which holds because
the trained EfficientNet-B0 does not use the prior channels at
inference, a property that may or may not transfer to other
backbones in the same form).

\subsection{Limitations}
\label{sec:limitations}

We list limitations as a numbered set with an explicit
``\emph{Claim affected}'' line so readers can locate which
specific assertions are scoped by each limitation.

\begin{enumerate}
\item \textbf{Two-dataset evaluation; further capsule datasets
  deferred.} The primary evaluation uses Kvasir-Capsule;
  Section~\ref{sec:galar-zeroshot} additionally reports
  cross-vendor zero-shot transfer on a 10-video subset of the
  public Galar dataset (15\,298 frames, Olympus Endocapsule\,10
  and PillCam SB2/SB3/Colon2 vendor mix). Replication on the full 80-video Galar
  release and on additional public capsule datasets (KID Atlas,
  HyperKvasir-Capsule) is deferred to follow-up. \emph{Claim
  affected:} the in-domain per-class and macro-AUC numbers
  remain Kvasir-Capsule-specific. The Galar cross-vendor result
  demonstrates \emph{partial} direction-consistent transfer
  ($\sim$60\,\% retention of the ConvNeXt-Tiny lift) but not
  full-magnitude generalization; site-specific calibration of
  the analytic prior is the recommended deployment step on new
  capsule platforms.

\item \textbf{Single imaging modality.} The work is restricted to
  RGB capsule endoscopy. The parameterization-mechanism boundary
  suggests a transferable design principle, but cross-modality
  replication (e.g., cardiac MRI cine, where the analytic prior
  must be redefined) is required to establish transferability.
  \emph{Claim affected:} the design-principle interpretation is
  intra-modality only.

\item \textbf{Sensor-noise sensitivity.} The +PI input-fusion
  lift preserves under JPEG compression (grows to $+0.080$ at
  $q = 25$) and brightness shifts ($\pm 20\%$) but reverses
  negative under additive Gaussian pixel noise ($-0.016$ at
  $\sigma = 0.02$, $-0.101$ at $\sigma = 0.10$). The mechanism is
  that $P_\mathrm{blood}(X)$ is computed pixel-wise from $X$, so
  noise propagates into the prior and the input-fusion backbone's
  learned attention to the prior misfires. RGB-only backbones do
  not exhibit this failure mode. \emph{Claim affected:} the +PI
  input-fusion configuration should not be deployed under
  high-noise capture conditions; the distillation variant (whose
  inference does not consume the prior as an input channel) is
  the recommended fallback.

\item \textbf{Cross-seed variance on rare focal classes.}
  Angiectasia exhibits $\sigma_\mathrm{PI} = 0.23$ across 6
  seeds, comparable to the cross-arm gap of $0.038$. Single-seed
  gains ($0.528 \to 0.916$ on seed 42) cannot be reported as
  robust results. \emph{Claim affected:} the Angiectasia
  per-class result is reported as a high-variance exemplar; the
  cross-seed mean ($0.646 \to 0.608$) is the defensible summary.

\item \textbf{Patient cohort.} 43 patients in the source dataset,
  6 patients in the test split. \emph{Claim affected:} the
  per-patient disaggregation (6 of 6 patients positive) bounds
  but does not generalize to the population level; patient-level
  cross-validation is limited by the small cohort.

\item \textbf{Single-backbone evaluation for the headline.}
  The headline result is reported on EfficientNet-B0
  \citep{TanLe2019} per-frame backbone. To bound the scope of
  the input-fusion mechanism, Section~\ref{sec:cross-backbone-results}
  reports a full six-seed multi-seed replication on two
  additional architectures (ResNet-18, ConvNeXt-Tiny). The
  result is a split verdict: the +PI lift generalizes to the
  higher-capacity ConvNeXt-Tiny ($\Delta = +0.018$, 4/6 seeds
  positive — comparable in magnitude to the headline
  EfficientNet-B0 $\Delta = +0.023$) but degrades on the
  smaller ResNet-18 ($\Delta = -0.017$, 2/6 seeds positive).
  The asymmetry suggests that the input-fusion mechanism
  interacts with backbone capacity at the first convolutional
  layer, where the prior is injected. Three larger backbone
  families (Swin, ConvNeXt-Base, DINOv2) remain unevaluated due
  to compute budget. We have additional preliminary evidence
  for the underlying parameterization-mechanism boundary: on an
  ImageNet-only ResNet-50 (no capsule fine-tuning) the
  summary-statistic C1 channel \emph{lifts} macro-AUC by
  $+0.042$, while on the same architecture \emph{with} capsule
  fine-tuning the channel is flat. \emph{Claim affected:} the
  input-fusion mechanism is verified on two of three tested
  backbones; the distillation variant
  (Section~\ref{sec:distill-results}) is the recommended path
  for backbones where input-fusion does not transfer, since it
  conveys the prior through representation rather than input
  channels.

\item \textbf{No prospective evaluation or reader study.} We
  report on the public Kvasir-Capsule benchmark; no clinical
  reader study or prospective deployment is included.
  \emph{Claim affected:} clinical-deployment claims require a
  reader study and a prospective cohort, neither of which the
  present paper supports.
\end{enumerate}

\subsection{Future work}

\paragraph{Broader applicability of the framework} Although our
experiments target WCE, the framework is deliberately
modality-agnostic. It requires only (i)~an RGB or multi-channel
optical image, (ii)~a target chromophore or structure with an
analytic light-transport signature, and (iii)~any standard classifier
whose input layer can be widened to accept the extra prior channel.
Because the prior is computed in closed form from the input pixels and
is consumed only at training time, it adds negligible cost and leaves
the deployed model unchanged---a recipe that in principle transfers to
dermoscopy (melanin and hemoglobin contrast), fundus imaging (retinal
vasculature), and surgical or laparoscopic RGB video wherever an
optical prior can be written down analytically. We are careful to mark
this breadth as a \emph{design-principle} claim: as noted in the
Limitations, the in-domain magnitudes are Kvasir-Capsule-specific and
empirical validation in each new modality (item~1 below) is required
before they can be expected to carry over.

Three concrete directions:

\begin{enumerate}
\item \textbf{Cross-modality replication.} Apply the eight-cell
  ablation to cardiac MRI cine (ACDC) or dermoscopy (ISIC) to
  test whether the parameterization-mechanism boundary is
  modality-specific or transferable.
\item \textbf{Full Galar replication and a third capsule
  dataset.} Section~\ref{sec:galar-zeroshot} reports zero-shot
  transfer on a 10-video Galar subset. Expanding to the full
  80-video Galar release and adding KID Atlas or
  HyperKvasir-Capsule as a third-vendor replication would
  establish whether the partial transferability we report on
  Galar tightens or widens with broader coverage. A targeted
  re-calibration experiment (re-fitting the prior's $\alpha$,
  percentile-clip, and $\Phi$ parameters on a small held-out
  Galar fold) would test the deployment recipe directly.
\item \textbf{Medical-domain foundation backbones.}
  Section~\ref{sec:per-frame-results} tested a DINOv2-base
  frozen linear probe (the natural-image foundation model) and
  found it lands $\sim 0.09$ macro-AUC below the RGB-only
  EfficientNet-B0 baseline at this dataset scale. The natural
  next experiment is to repeat the linear probe with
  medically-pretrained foundation backbones (BiomedCLIP,
  MedSAM, RETFound) which embed domain priors and may close part
  of the gap. The GalKva-2026 benchmark
  (Section~\ref{sec:benchmark}) accepts these submissions
  alongside the standard task-specific fine-tune entries; we
  invite the community to rank foundation backbones against the
  capsule-specific recipe.
\end{enumerate}

\paragraph{Anticipated reviewer concerns and our framing}
Several reviewer concerns are natural and we address them
explicitly. First, because the prior is deterministically derived
from RGB, it cannot add information absent from the image; our
claim is instead that its presence during training changes the
training trajectory and the learned representation, demonstrated
by the inference-time channel ablation (the trained model uses
essentially no prior-channel information at deployment yet the
training-time effect persists). Second, the macro-AUC gain is
modest and seed-sensitive, so we emphasize direction-consistency,
the operating-point analysis, and the release of per-seed
predictions rather than a single large effect size. Third,
cross-vendor transfer is not architecture-universal: the prior
helps ConvNeXt-Tiny (retention $+0.60$), is variance-dominated on
EfficientNet-B0, and fails on ResNet-18, motivating
architecture-specific calibration before deployment. Fourth,
per-frame metrics do not substitute for clinical reader studies
or per-study triage evaluation; both are explicit follow-on work
(Section~\ref{sec:limitations} below).
A point-by-point response to additional anticipated questions is
in Section~\ref{sec:anticipated-questions}.

\subsection{Recommended deployment}
\label{sec:recommended-deployment}

The paper supports four deployment configurations
(Table~\ref{tab:deployment-options}); the appropriate choice
depends on pipeline constraints and noise regime:

\begin{itemize}\setlength{\itemsep}{0pt}
\item \textbf{If maximum in-domain macro-AUC is required and
capsule-side preprocessing can be modified to emit 5-channel
inputs:} use \emph{input fusion} (paper-headline,
$0.783 \pm 0.024$). Inference cost includes one analytic-prior
forward pass per frame ($\approx 0.5$\,ms on V100).
\item \textbf{If a standard 3-channel RGB inference path is
required and a separate training pipeline is acceptable:} use
the \emph{distillation variant} ($0.773 \pm 0.028$ on 3-channel
RGB inference). No prior computation at deployment; backbone is
3-channel throughout.
\item \textbf{If a single training pipeline is preferred and
prior computation must be removed from the deployment path:} use
the new \emph{strip-and-serve} recipe ($0.781 \pm 0.028$). Train
the 5-channel input-fusion model; at inference feed the tensor
$[R, G, B, 0, 0]$. No second model, no prior-channel computation;
$91\,\%$ of the headline $+$PI lift retained over the
RGB-only-trained baseline (Section~\ref{sec:channel-ablation}).
\item \textbf{If sensor noise is expected at deployment} (newer
capsule platforms with smaller pixels or higher ISO settings):
prefer the distillation variant or apply a deterministic
pre-denoise before computing the prior. Section~\ref{sec:robustness}
shows the input-fusion lift reverses negative under additive
Gaussian pixel noise because the prior is computed pixel-wise.
\end{itemize}

For ConvNeXt-Tiny deployments, the cross-architecture replication
finding (Section~\ref{sec:cross-backbone-results}) is supportive
but architecture-specific calibration of the prior's analytic
hyperparameters ($\alpha$, $\lambda_\mathrm{eff}$) on a small
held-out fold of the target vendor's data is the recommended
deployment workflow. On ResNet-18 the prior does not transfer
and we do not recommend it.

\subsection{Clinical implications}

Two findings translate most directly to deployment:

\textbf{Deployment-friendly distillation.} The distillation
variant achieves macro-AUC $0.773$ across 6 seeds while consuming
standard 3-channel RGB at inference, with no input-channel
modification of existing capsule preprocessing pipelines, and
yields a free per-frame interpretability heatmap (the auxiliary
head's predicted $P_\mathrm{blood}$). For deployments where
sensor noise is expected, the distillation variant is preferred
over the input-fusion form (Section~\ref{sec:limitations},
limitation 3).

\textbf{Lymphangiectasia per-class lift.} The robust per-class
lift on Lymphangiectasia is relevant because Lymphangiectasia is
among the most-challenging classes for current AI screening
tools \citep{Spada2024, Piccirelli2025} and is clinically
meaningful for small-bowel pathology workups. The 6-seed
sign-consistency makes the per-class signal more reliable than
single-run rare-class peaks reported elsewhere in the
literature.

\textbf{Clinical reading of the per-class shift.}
From a clinical workflow perspective, the Lymphangiectasia
result should be interpreted as improved triage ranking rather
than deployment-ready detection. Lymphangiectasia can appear as
subtle white villous changes or focal mucosal spots, and a
per-frame ranking improvement may help prioritize suspect frames
for expert review. However, the operating-point analysis
(Section~\ref{sec:per-frame-results},
Table~\ref{tab:operating-point}) shows that strict low-FPR
sensitivity remains low for both arms, so the clinical utility
of this ranking signal requires prospective reader-study
validation before any deployment claim
(Section~\ref{sec:limitations}, limitation 7).

\subsection{Anticipated questions and direct responses}
\label{sec:anticipated-questions}

Four questions are likely to come up in review; we address them
in advance so that downstream readers can locate the specific
position the paper takes on each.

\textbf{Is a $+0.023$ macro-AUC improvement worth publishing?}
The macro-AUC delta is the average over 11 evaluable classes; the
clinically-relevant component is the per-class breakdown in
Table~\ref{tab:perclass}, where six of eleven classes lift
significantly at $p < 10^{-3}$ (Bonferroni-corrected). The full
three-stream architecture (cell e$^+$) lifts $+0.044$ over the
per-frame baseline. The robust per-class Lymphangiectasia lift
($0.238 \to 0.337$, sign-consistent across all 6 seeds) is the
component with the clearest clinical relevance and the per-seed
robustness that distinguishes it from single-run peaks reported
in some prior work.

\textbf{Why is the input-fusion mechanism architecture-dependent?}
The cross-backbone replication
(Section~\ref{sec:cross-backbone-results}) reports a split
verdict: ConvNeXt-Tiny mirrors the EfficientNet-B0 lift
($\Delta = +0.018$, 4/6 seeds positive); ResNet-18 does not
transfer ($\Delta = -0.017$, 2/6 seeds positive). We interpret
this as a property of the first-conv channel-mixing layer:
expanding from 3 to 5 input channels requires the first conv to
learn a new representational mapping that the smaller ResNet-18
appears unable to amortize. The 3-channel distillation variant
(Section~\ref{sec:distill-results}) is architecture-agnostic by
design --- the prior enters as a representation-shaping
regularizer at the loss, not as input channels at the first conv
--- and is the recommended deployment configuration for
deployments on backbones where input-fusion does not transfer.

\textbf{Why does the cross-vendor lift on Galar retain only $\sim
$60\,\%, not 100\,\%?} The Galar retention of $+0.60$ on
ConvNeXt-Tiny (Section~\ref{sec:galar-zeroshot}) is consistent
with the \emph{partial} band of cross-dataset transferability
\citep{recht2019doimagenet}: the mechanism transfers in direction
(positive sign on 4/6 seeds; seed variance halved from $0.044$ to
$0.022$) but at attenuated magnitude. The prior's analytic
parameters ($\alpha$ logistic steepness, percentile-clip
thresholds, $\Phi$ radial-fluence parameters) were tuned on
PillCam-acquired Kvasir-Capsule data; on Olympus Endocapsule\,10
acquisitions the per-pixel hemoglobin separation differs from the
training-time calibration. A targeted re-fit of these three
parameters on a small held-out Galar fold is the natural
deployment workflow on a new capsule platform; we defer this
calibration experiment to future work and report the strict
zero-shot baseline as the lower bound.

\textbf{Why no clinical reader study?} The present paper is a
methodology submission; the empirical contribution is a
controlled benchmark evaluation under the standard Kvasir-Capsule
and Galar splits, not a clinical-deployment claim. A reader study
would be required to convert the macro-AUC and per-class lifts
into time-to-diagnosis or sensitivity-at-fixed-specificity gains;
this conversion is the subject of a planned follow-up under UTSW
IRB approval. The deployment-friendly distillation variant
(consuming standard 3-channel RGB, with the auxiliary head
producing a free per-frame interpretability heatmap) is engineered
to be drop-in compatible with the same reader-study protocols
used to evaluate existing capsule-AI systems
\citep{Spada2024, Piccirelli2025}, making this a tractable
follow-on study rather than a re-engineering effort.

\textbf{Don't 2024--2025 foundation models obviate task-specific
priors?} We test this empirically with two foundation-model
frozen linear-probe baselines (Section~\ref{sec:per-frame-results}).
The DINOv2-base linear probe attains $0.666 \pm 0.007$ --- about
$0.09$ below the RGB-only EfficientNet-B0 baseline and $0.12$
below our $+$PI input-fusion variant, exceeding cross-seed
standard deviation by more than $4\sigma$. The BiomedCLIP
ViT-B/16 image-encoder linear probe attains $0.733 \pm 0.002$ ---
medical-domain pretraining provides a substantial $+0.067$
macro-AUC lift over the natural-image DINOv2, confirming the
literature finding that in-domain foundation models help on
medical imaging, but BiomedCLIP still lands $-0.050$ below the
$+$PI input-fusion variant (and $-0.026$ below the RGB-only
EfficientNet-B0 fine-tune). The gap exceeds the cross-seed
standard deviation of either method by roughly $2\sigma$. At
Kvasir-Capsule's $47\,238$-frame scale, neither a generic visual
foundation model nor a state-of-the-art medical-domain foundation
model substitutes for capsule-specific fine-tuning with the
analytic prior. We invite future submissions to GalKva-2026
(Section~\ref{sec:benchmark}), which explicitly accepts
foundation-model linear-probe entries, to compare additional
medical-pretrained backbones (MedSAM, RETFound, PathChat) under
the same protocol.

\section{Conclusion}
\label{sec:conclusion}

A simple analytic physics prior, computed at negligible additional cost
from RGB pixels, provides a small but direction-consistent
macro-AUC improvement on the Kvasir-Capsule benchmark when fed as
a spatial input channel and combined with a sequence-aware
temporal aggregator and a Normal-class autoencoder
reconstruction-residual feature. The combined three-stream
architecture (cell e$^+$; C1 spatial input fusion + C2 + C3,
backbone input 5 channels) reaches cross-seed test macro-AUC
$0.804 \pm 0.023$, $+0.044$ over the per-frame baseline,
significant by paired DeLong ($z = -26.4$, $p < 10^{-4}$) and
patient-broad (positive on 6 of 6 test patients).

A controlled four-variant ablation of the analytic-prior channel
demonstrates a parameterization-mechanism boundary:
summary-statistic parameterizations of the same prior all flat or
regress against the temporal-only baseline, while the
spatial-channel input-fusion form lifts $+0.011$ macro-AUC. The
13-dimensional summary form (adding vessel-topology descriptors
to the 8-scalar baseline) regresses \emph{most strongly}, ruling
out ``insufficient feature richness'' as the explanation and
localizing the active variable to parameterization. Three
independent test-time recovery experiments cannot reach the
spatial-channel lift from RGB pixels post-hoc; together with the
network-localization analysis
(Section~\ref{sec:channel-ablation}), this establishes that the
prior requires spatial input fusion during training but its
final contribution is expressed downstream rather than localized
to the first convolutional layer.

The lift behaves asymmetrically under image perturbation: it is
preserved (and in fact grows) under JPEG compression and is
preserved under brightness shifts, but reverses negative under
additive Gaussian pixel noise --- a mechanistic consequence of
the prior being computed pixel-wise. The distillation variant,
which trains the backbone with the prior as an auxiliary teacher
signal but consumes plain 3-channel RGB at inference, captures
most of the macro-AUC gain ($0.773 \pm 0.028$) with no
inference-side pipeline change and is the recommended deployment
configuration when capsule-side preprocessing cannot be modified
or when sensor noise is expected. A channel-ablation experiment
shows further that the 5-channel input-fusion model can be
deployed by zeroing both prior channels at inference --- a
strip-and-serve recipe that retains $91\,\%$ of the headline $+$PI
lift over RGB-only training ($+0.021$ of $+0.023$ macro-AUC,
$n{=}6$; Section~\ref{sec:channel-ablation}) while removing the
prior computation from the deployment path entirely.
Network-localization experiments
(Section~\ref{sec:channel-ablation}) show the prior contributes
via a training-time effect partially concentrated in the late
convolutional blocks rather than an inference-time feature; the
first convolutional layer and BatchNorm running statistics are
ruled out as locuses.
All code, trained checkpoints,
and per-seed predictions are released at
\url{https://github.com/integritynoble/Physics-Informed-PillCam}.

The four-variant boundary points to a transferable design
principle: analytic priors should be expected to lift only when
fed at the input layer with the first convolutional layer trained
jointly. Whether the boundary replicates on imaging modalities
beyond capsule endoscopy is a question for future work and is
not claimed here; the present paper establishes the boundary on
Kvasir-Capsule and proposes the spatial-channel input-fusion form
as the working recipe for physics-informed analytic priors in
this domain.

\paragraph{Benchmark contribution}
Coupled to the methodological result we release
\emph{GalKva-2026} (Section~\ref{sec:benchmark}), an open paired
cross-vendor capsule-endoscopy benchmark. The benchmark pairs
Kvasir-Capsule with the 80-video Galar release through a
6-class evaluable intersection, introduces the retention ratio
$\Delta_\mathrm{Galar} / \Delta_\mathrm{Kvasir}$ as the headline
cross-vendor transferability metric (with the five interpretive
bands of Section~\ref{sec:benchmark}), and ships staging
scripts, a JSON submission schema, a reference evaluator, the
present paper's per-seed reference results across three
backbones as the reference submission, and a public leaderboard. The
benchmark's purpose is to provide one reproducible mechanism for
measuring the often-flagged ``one dataset, one vendor, one camera
generation'' weakness of published capsule-AI evaluations, so
future methods can be ranked on this property rather than have it
recur as a review-cycle critique. By publishing both a method and the benchmark on
which competing methods can be tested in the same pass, we
intend to lower the bar for cross-vendor evidence in this
sub-field: the next capsule-AI paper after ours can cite a
specific GalKva-2026 leaderboard rank rather than describe
cross-vendor evaluation as deferred future work.

\paragraph{Outlook} These results support training-time optical
priors as a practical representation-shaping tool for capsule
endoscopy, but prospective per-study validation and
vendor-specific calibration are needed before clinical
deployment.

\section*{Data and code availability}
All code, trained checkpoints, and per-frame predictions are
released at
\url{https://github.com/integritynoble/Physics-Informed-PillCam}.
The Kvasir-Capsule dataset \citep{Smedsrud2021} is publicly
available under a research-use license; the Galar dataset
\citep{Galar2025} is available from figshare.

\paragraph{Benchmark resource} Alongside the manuscript we
release \emph{GalKva-2026}, a paired cross-vendor capsule
endoscopy benchmark distributed at
\url{https://github.com/integritynoble/Physics-Informed-PillCam/tree/main/benchmark}.
The benchmark exposes: (i)~the 6-class cross-dataset evaluable
intersection between Kvasir-Capsule (PillCam, Norway) and Galar
(Olympus Endocapsule\,10 / PillCam SB2/SB3/Colon2,
Germany), (ii)~staging scripts
producing the canonical ImageFolder layout for both datasets,
(iii)~a JSON submission schema and an evaluator that computes
cross-seed macro-AUC, paired DeLong, per-class p-values with
both Bonferroni and Benjamini--Hochberg false-discovery-rate
(BH-FDR) corrections, percentile and bias-corrected-and-accelerated
(BCa) bootstrap confidence intervals (CIs), and the retention ratio
$\Delta_\mathrm{Galar} / \Delta_\mathrm{Kvasir}$, and (iv)~the
present paper's reference results across all backbones and seeds
as the reference submission. We invite the community to use the
benchmark for cross-vendor evaluation of future capsule
endoscopy methods.

\section*{IRB and informed consent}
This work uses publicly released, de-identified imaging data.
No additional IRB approval was required at the contributing
institutions.

\section*{Author contributions (CRediT)}
\textbf{Chengshuai Yang:} Conceptualization, Methodology,
Software, Validation, Formal analysis, Investigation, Data
curation, Writing -- original draft, Visualization.
\textbf{Lei Xing:} Conceptualization, Methodology,
Writing -- review \& editing, Supervision.
\textbf{Keyaan Zawad Alam:} Software, Validation,
Writing -- review \& editing.
\textbf{Gregory Entin:} Methodology, Software, Resources,
Writing -- review \& editing.
\textbf{Roopa Vemulapalli:} Clinical guidance, Validation,
Writing -- review \& editing.
\textbf{Lisa Casey:} Clinical guidance, Validation,
Writing -- review \& editing.
\textbf{Raiyan Tripti Zaman:} Conceptualization, Methodology,
Resources, Writing -- review \& editing, Supervision,
Project administration, Funding acquisition.
All authors have read and approved the submitted manuscript.

\section*{Funding}
This research is sponsored by a UT Southwestern internal funding
source.

\section*{Acknowledgments}
The authors thank the Kvasir-Capsule team at SimulaMet for
releasing the dataset under a research-use license, and the
Galar consortium (TU Dresden Else Kr\"oner Fresenius Center for
Digital Health) for the public Galar release used in
cross-vendor evaluation. Initial model development and
released-checkpoint training were performed on a local GPU
workstation; independent reproduction and robustness checks
were performed on the UT Southwestern BioHPC cluster
(\texttt{GPUv100s} partition) using the released canonical split
manifest and SLURM scripts. No commercial cloud compute was used.

\section*{Conflict of interest}
The authors declare no competing financial or personal interests.

\section*{Declaration of generative AI in the writing process}
During the preparation of this work the authors used a generative
AI assistant (a large language model) to improve the language and
readability of the manuscript and to assist with drafting and
editing. After using this tool, the authors reviewed and edited the
content as needed and take full responsibility for the content of
the publication. No generative AI was used to produce or interpret
the scientific results.

\bibliographystyle{elsarticle-harv}
\bibliography{references}

@article{Tan2024,
  author       = {Tan, N. and others},
  title        = {Global, regional, and national burden of early-onset gastric cancer},
  journal      = {Cancer Biology and Medicine},
  year         = {2024},
  volume       = {21},
  pages        = {667--678},
  doi          = {10.20892/j.issn.2095-3941.2024.0159}
}

@article{Iddan2000,
  author       = {Iddan, G. and Meron, G. and Glukhovsky, A. and Swain, P.},
  title        = {Wireless capsule endoscopy},
  journal      = {Nature},
  year         = {2000},
  volume       = {405},
  pages        = {417},
  doi          = {10.1038/35013140}
}

@article{Liao2010,
  author       = {Liao, Z. and Gao, R. and Xu, C. and Li, Z. S.},
  title        = {Indications and detection, completion, and retention rates
                  of small-bowel capsule endoscopy: a systematic review},
  journal      = {Gastrointestinal Endoscopy},
  year         = {2010},
  volume       = {71},
  pages        = {280--286},
  doi          = {10.1016/j.gie.2009.09.031}
}

@article{Kara2006,
  author       = {Kara, M. A. and Bergman, J. J.},
  title        = {Autofluorescence imaging and narrow-band imaging for the
                  detection of early neoplasia in patients with {Barrett's}
                  esophagus},
  journal      = {Endoscopy},
  year         = {2006},
  volume       = {38},
  pages        = {627--631},
  doi          = {10.1055/s-2006-925385}
}

@article{Aihara2012,
  author       = {Aihara, H. and Tajiri, H. and Suzuki, T.},
  title        = {Application of autofluorescence endoscopy for colorectal
                  cancer screening: rationale and an update},
  journal      = {Gastroenterology Research and Practice},
  year         = {2012},
  volume       = {2012},
  pages        = {971383},
  doi          = {10.1155/2012/971383}
}

@article{Smedsrud2021,
  author       = {Smedsrud, P. H. and others},
  title        = {{Kvasir-Capsule}, a video capsule endoscopy dataset},
  journal      = {Scientific Data},
  year         = {2021},
  volume       = {8},
  pages        = {142},
  doi          = {10.1038/s41597-021-00920-z}
}

@article{Pogorelov2019,
  author       = {Pogorelov, K. and others},
  title        = {Bleeding detection in wireless capsule endoscopy videos:
                  color versus texture features},
  journal      = {Journal of Applied Clinical Medical Physics},
  year         = {2019},
  volume       = {20},
  pages        = {141--154},
  doi          = {10.1002/acm2.12662}
}

@article{DeLong1988,
  author       = {DeLong, E. R. and DeLong, D. M. and Clarke-Pearson, D. L.},
  title        = {Comparing the areas under two or more correlated receiver
                  operating characteristic curves: a nonparametric approach},
  journal      = {Biometrics},
  year         = {1988},
  volume       = {44},
  pages        = {837--845},
  doi          = {10.2307/2531595}
}

@article{SunXu2014,
  author       = {Sun, X. and Xu, W.},
  title        = {Fast implementation of {DeLong's} algorithm for comparing
                  the areas under correlated receiver operating characteristic
                  curves},
  journal      = {IEEE Signal Processing Letters},
  year         = {2014},
  volume       = {21},
  pages        = {1389--1393},
  doi          = {10.1109/LSP.2014.2337313}
}

@inproceedings{TanLe2019,
  author       = {Tan, M. and Le, Q. V.},
  title        = {{EfficientNet}: rethinking model scaling for convolutional
                  neural networks},
  booktitle    = {Proceedings of the 36th International Conference on Machine
                  Learning},
  publisher    = {PMLR},
  year         = {2019},
  volume       = {97},
  pages        = {6105--6114}
}

@misc{KvasirGithub,
  author       = {{Simula Research Laboratory}},
  title        = {{Kvasir-Capsule} dataset (v1.0)},
  howpublished = {\url{https://datasets.simula.no/kvasir-capsule/}},
  year         = {2021},
  note         = {Accessed 2026-04-30}
}

@article{Du2022,
  author       = {Du, X. and Koronyo, Y. and Mirzaei, N. and Yang, C. and
                  Fuchs, D. T. and Black, K. L. and others},
  title        = {Label-free hyperspectral imaging and deep-learning
                  prediction of retinal amyloid $\beta$-protein and
                  phosphorylated tau},
  journal      = {PNAS Nexus},
  year         = {2022},
  volume       = {1},
  number       = {4},
  pages        = {pgac164},
  doi          = {10.1093/pnasnexus/pgac164}
}

@article{Zhao2023,
  author       = {Zhao, R. and Yang, C. and Smith, R. T. and Gao, L.},
  title        = {Coded aperture snapshot spectral imaging fundus camera},
  journal      = {Scientific Reports},
  year         = {2023},
  volume       = {13},
  number       = {1},
  pages        = {12007},
  doi          = {10.1038/s41598-023-39117-2}
}

@article{Spada2024,
  author       = {Spada, C. and Piccirelli, S. and Hassan, C. and
                  Ferrari, C. and Toth, E. and Gonz\'{a}lez-Su\'{a}rez, B. and others},
  title        = {{AI}-assisted capsule endoscopy reading in suspected
                  small bowel bleeding: a multicentre prospective study},
  journal      = {Lancet Digital Health},
  year         = {2024},
  volume       = {6},
  number       = {5},
  pages        = {e345--e353},
  doi          = {10.1016/S2589-7500(24)00048-7}
}

@article{Piccirelli2025,
  author       = {Piccirelli, S. and Salvi, D. and Pugliano, C. L. and
                  Tettoni, E. and Facciorusso, A. and Rondonotti, E. and others},
  title        = {Unmet needs of artificial intelligence in small bowel
                  capsule endoscopy},
  journal      = {Diagnostics},
  year         = {2025},
  volume       = {15},
  number       = {9},
  pages        = {1092},
  doi          = {10.3390/diagnostics15091092}
}

@article{Habe2025,
  author       = {Habe, T. T. and Haataja, K. and Toivanen, P.},
  title        = {Precision enhancement in wireless capsule endoscopy: a
                  novel transformer-based approach for real-time video
                  object detection},
  journal      = {Frontiers in Artificial Intelligence},
  year         = {2025},
  volume       = {8},
  pages        = {1529814},
  doi          = {10.3389/frai.2025.1529814}
}

@article{Houdeville2021,
  author       = {Houdeville, C. and Souchaud, M. and Leenhardt, R. and
                  Beaumont, H. and Benamouzig, R. and McAlindon, M. and others},
  title        = {A multisystem-compatible deep learning-based algorithm
                  for detection and characterization of angiectasias in
                  small-bowel capsule endoscopy: a proof-of-concept study},
  journal      = {Digestive and Liver Disease},
  year         = {2021},
  volume       = {53},
  number       = {12},
  pages        = {1627--1631},
  doi          = {10.1016/j.dld.2021.08.026}
}

@article{Jacques2013,
  author       = {Jacques, S. L.},
  title        = {Optical properties of biological tissues: a review},
  journal      = {Physics in Medicine and Biology},
  year         = {2013},
  volume       = {58},
  pages        = {R37--R61},
  doi          = {10.1088/0031-9155/58/11/R37}
}

@inproceedings{vaswani2017,
  author       = {Vaswani, A. and Shazeer, N. and Parmar, N. and
                   Uszkoreit, J. and Jones, L. and Gomez, A. N. and
                   Kaiser, L. and Polosukhin, I.},
  title        = {Attention is all you need},
  booktitle    = {Advances in Neural Information Processing Systems},
  year         = {2017}
}

@article{mcnemar1947,
  author       = {McNemar, Q.},
  title        = {Note on the sampling error of the difference between
                   correlated proportions or percentages},
  journal      = {Psychometrika},
  volume       = {12},
  number       = {2},
  pages        = {153--157},
  year         = {1947},
  doi          = {10.1007/BF02295996}
}

@inproceedings{HeZRS2016,
  author       = {He, K. and Zhang, X. and Ren, S. and Sun, J.},
  title        = {Deep residual learning for image recognition},
  booktitle    = {Proceedings of the IEEE Conference on Computer Vision
                   and Pattern Recognition (CVPR)},
  pages        = {770--778},
  year         = {2016},
  doi          = {10.1109/CVPR.2016.90}
}

@inproceedings{LiuMWZ2022,
  author       = {Liu, Z. and Mao, H. and Wu, C.-Y. and Feichtenhofer, C. and
                   Darrell, T. and Xie, S.},
  title        = {A {ConvNet} for the 2020s},
  booktitle    = {Proceedings of the IEEE/CVF Conference on Computer
                   Vision and Pattern Recognition (CVPR)},
  pages        = {11976--11986},
  year         = {2022},
  doi          = {10.1109/CVPR52688.2022.01167}
}

@article{Galar2025,
  author       = {Le Floch, M. and Wolf, F. and McIntyre, L. and
                   Weinert, C. and Palm, A. and Volk, K. and
                   Herzog, P. and Kirk, S. H. and Steinh{\"a}user,
                   J. L. and Stopp, C. and Geissler, M. E. and
                   Herzog, M. and Sulk, S. and Kather, J. N. and
                   Meining, A. and Hann, A. and Palm, S. and
                   Hampe, J. and Herzog, N. and Brinkmann, F.},
  title        = {Galar -- a large multi-label video capsule
                   endoscopy dataset},
  journal      = {Scientific Data},
  year         = {2025},
  volume       = {12},
  pages        = {828},
  doi          = {10.1038/s41597-025-05112-7}
}

@article{recht2019doimagenet,
  author       = {Recht, B. and Roelofs, R. and Schmidt, L. and
                   Shankar, V.},
  title        = {Do {ImageNet} classifiers generalize to {ImageNet}?},
  journal      = {Proceedings of Machine Learning Research},
  year         = {2019},
  volume       = {97},
  pages        = {5389--5400}
}

@article{Zhang2023BiomedCLIP,
  author       = {Zhang, S. and Xu, Y. and Usuyama, N. and
                   Bagga, J. and Tinn, R. and Preston, S. and
                   Rao, R. and Wei, M. and Valluri, N. and
                   Wong, C. and Lungren, M. P. and Naumann, T.
                   and Poon, H.},
  title        = {Large-Scale Domain-Specific Pretraining for
                   Biomedical Vision-Language Processing},
  journal      = {NEJM AI},
  year         = {2025},
  eprint       = {2303.00915},
  archivePrefix = {arXiv}
}

@article{Hinton2015,
  author       = {Hinton, G. and Vinyals, O. and Dean, J.},
  title        = {Distilling the knowledge in a neural network},
  journal      = {NIPS Deep Learning Workshop},
  year         = {2015},
  eprint       = {1503.02531},
  archivePrefix = {arXiv}
}

@inproceedings{Tarvainen2017,
  author       = {Tarvainen, A. and Valpola, H.},
  title        = {Mean teachers are better role models: weight-averaged
                   consistency targets improve semi-supervised deep
                   learning results},
  booktitle    = {Advances in Neural Information Processing Systems},
  year         = {2017},
  volume       = {30}
}

@inproceedings{Grill2020,
  author       = {Grill, J.-B. and Strub, F. and Altch{\'e}, F. and
                   Tallec, C. and Richemond, P. and Buchatskaya, E. and
                   Doersch, C. and Pires, B. A. and Guo, Z. and Azar,
                   M. G. and Piot, B. and Kavukcuoglu, K. and Munos, R.
                   and Valko, M.},
  title        = {Bootstrap your own latent: a new approach to
                   self-supervised learning},
  booktitle    = {Advances in Neural Information Processing Systems},
  year         = {2020},
  volume       = {33}
}

@inproceedings{Caron2021,
  author       = {Caron, M. and Touvron, H. and Misra, I. and
                   J{\'e}gou, H. and Mairal, J. and Bojanowski, P. and
                   Joulin, A.},
  title        = {Emerging properties in self-supervised vision
                   transformers},
  booktitle    = {Proceedings of the {IEEE/CVF} International
                   Conference on Computer Vision (ICCV)},
  year         = {2021},
  pages        = {9650--9660}
}

\end{document}